\documentclass[lettersize,journal]{IEEEtran}
\usepackage{amsmath}
\usepackage{amssymb}
\usepackage{amssymb}
\usepackage[shortlabels]{enumitem}
\usepackage{amsmath,amsthm}
\usepackage{booktabs}
\usepackage{textcomp}
\usepackage{multirow}
\usepackage{colortbl}
\usepackage{color}
\usepackage{xcolor}
\usepackage{graphicx}
\usepackage{subfigure}
\usepackage{algpseudocode}
\usepackage{color}
\usepackage{graphicx}
\usepackage[strings]{underscore}
\usepackage{algpseudocode}
\usepackage{caption}
\usepackage{threeparttable}
\usepackage{epstopdf}
\usepackage{url}
\usepackage{mathrsfs}
\usepackage{gensymb}

\makeatletter
\newif\if@restonecol
\makeatother

\usepackage[linesnumbered,ruled,vlined]{algorithm2e}%[ruled,vlined]{
\usepackage{algpseudocode}
\usepackage{amsmath}
\usepackage{doi}

\usepackage{bm}
\usepackage{cite}

\begin{document}

\title{Traj-LLM: A New Exploration for Empowering Trajectory Prediction with Pre-trained Large Language Models}
  
 % Empower Trajectory Prediction by Large Language Models Preliminary explorations an initial attempt
 % Bootstapping Trajectory Prediction with Large Language Models
% \author{{}

% \thanks{Manuscript received xxx; revised xxx; accepted xxx.

% }% <-this % stops a space}

% }

\author{{Zhengxing Lan, Hongbo Li, Lingshan Liu, Bo Fan, Yisheng Lv, Senior Member, IEEE, \\
Yilong Ren, Member, IEEE, Zhiyong Cui, Member, IEEE}

\thanks{Manuscript received xxx; revised xxx; accepted xxx. This work was supported by the National Key Research and Development Project of China under Grant 2022YFB4300400, the National Natural Science Foundation of China under Grant 52202378, the Open Research Project Program of the State Key Laboratory of Internet of Things for Smart City under Grant SKL-IoTSC(UM)-2021-2023/ORP/GA08/2022, the Chunhui Collaboration Project Program of the Ministry of Education of China under Grant 202200650, and the Youth Talent Support Program of Beihang University under Grant YWF-22-L-1239. (Corresponding author: Yilong Ren, Zhiyong Cui.)

Zhengxing Lan is with the School of Transportation Science and Engineering, Beihang University, Beijing 100191, China, and also with the State Key Lab of Intelligent Transportation System, Beijing 100191, China.

Hongbo Li is with the School of Transportation Science and Engineering, Beihang University, Beijing 100191, China.

Lingshan Liu is with the Research Institute for Frontier Science, Beihang University, Beijing 100191, China, and also with the State Key Lab of Intelligent Transportation System, Beijing 100191, China.

Bo Fan is with the Beijing Key Laboratory of Traffic Engineering, College of Metropolitan Transportation, Beijing University of Technology, Beijing 100124, China.

Yisheng Lv is with the State Key Laboratory of Multimodal Artificial Intelligence Systems, Institute of Automation, Chinese Academy of Sciences, Beijing 100190, China, and also with the School of Artificial Intelligence, University of Chinese Academy of Sciences, Beijing 100049.

Yilong Ren and Zhiyong Cui are with the School of Transportation Science and Engineering, Beihang University, Beijing 100191, China, also with the State Key Lab of Intelligent Transportation System, Beijing 100191, China, also with the Zhongguancun Laboratory, Beijing 100194, China, and also with the Beihang Hangzhou Innovation Institute Yuhang, Hangzhou 310023, China.

}% <-this % stops a space}
}



\maketitle

\begin{abstract}
     Predicting the future trajectories of dynamic traffic actors is a cornerstone task in autonomous driving. Though existing notable efforts have resulted in impressive performance improvements, a gap persists in scene cognitive and understanding of the complex traffic semantics. This paper proposes Traj-LLM, the first to investigate the potential of using Large Language Models (LLMs) without explicit prompt engineering to generate future motion from agents’ past/observed trajectories and scene semantics. Traj-LLM starts with sparse context joint coding to dissect the agent and scene features into a form that LLMs understand. On this basis, we innovatively explore LLMs' powerful comprehension abilities to capture a spectrum of high-level scene knowledge and interactive information. Emulating the human-like lane focus cognitive function and enhancing Traj-LLM's scene comprehension, we introduce lane-aware probabilistic learning powered by the pioneering Mamba module. Finally, a multi-modal Laplace decoder is designed to achieve scene-compliant multi-modal predictions. Extensive experiments manifest that Traj-LLM, fortified by LLMs' strong prior knowledge and understanding prowess, together with lane-aware probability learning, outstrips state-of-the-art methods across evaluation metrics. Moreover, the few-shot analysis further substantiates Traj-LLM's performance, wherein with just 50\% of the dataset, it outperforms the majority of benchmarks relying on complete data utilization. This study explores equipping the trajectory prediction task with advanced capabilities inherent in LLMs, furnishing a more universal and adaptable solution for forecasting agent motion in a new way. 
\end{abstract}


\section{Introduction}
\IEEEPARstart{P}{redicting} the trajectories of agents primarily reasons how the future might unfold based on the observed behaviors of road users. Serving as an essential bridge connecting the perception system upward and the planning system downward, it assumes a pivotal role in the domain of autonomous driving. However, owing to the inherent stochasticity of agent motion, imbuing autonomous vehicles (AVs) with the capacity to anticipate surrounding trajectories as proficiently as humans remains a formidable challenge.

Recent works have focused on novel deep network architectures, offering promising avenues for enhancing prediction efficacy. These endeavors, characterized by modeling temporal dependencies and spatial interactions, reframe the trajectory prediction task as a time-series forecasting problem. A significant number of them apply distinct modules to mine temporal dynamics along with spatial relationships, and further integrate trajectories into composite models that possess both spatial and temporal components,  allowing for an understanding of spatio-temporal characteristics \cite{wu2023multi}. Specifically, in encoding temporal interdependencies, a variety of classical networks, exemplified by Recurrent Neural Networks (RNNs) \cite{mo2021graph} and their derivatives (e.g., LSTM \cite{park2018sequence,liao2024bat}, GRU \cite{sheng2022graph,liao2024physics}, and Transformer \cite{geng2023dynamic}) commonly play pivotal roles. Furthermore, in capturing spatial interactions, numerous emerging techniques such as the pooling mechanism \cite{zuo2023trajectory} and Graph Convolutional Networks (GCNs) \cite{ren2024emsin}, elaborately integrate the influential information from surrounding vehicles. Though these advancements have resulted in impressive performance improvements, a notable gap persists in scene cognition and understanding. The complex real driving conditions call for prediction schemes with more powerful modeling to capture a spectrum of high-level scene knowledge and interactive information, just as skilled drivers would.

The latest developments in Large Language Models (LLMs) have created new avenues for addressing challenges related to trajectory prediction. LLMs have demonstrated the remarkable ability to mimic human-like comprehension and reasoning across a spectrum of tasks \cite{liu2021kg,ouyang2022training}. Leveraging this potential, researchers have extrapolated LLMs to the domain of AVs, encompassing tasks such as motion planning, perception, and decision-making \cite{chen2023driving,wu2023language,wen2023dilu}. Notably, Bae et al. \cite{bae2024can} and Chib et al. \cite{chib2024lg} have explored the application of LLMs in pedestrian trajectory prediction, representing pioneering efforts in this direction. Their approach involves carefully crafting appropriate text prompts to harness the reasoning abilities of LLMs, thereby shifting the paradigm toward prompt-based predictions. However, crafting clear, systematically structured, and effective prompts poses challenges, necessitating meticulous attention to diverse facets such as query and response formulation \cite{xue2023promptcast}. Such diversity and uncertainty may introduce modeling bias, and prompt-based schemes cannot be easily incorporated into existing prediction methods. Indeed, another beauty of LLMs is to provide high-level connotations and extensive knowledge embedded within large pre-trained models \cite{radford2021learning}. The emergence phenomenon of LLMs has expanded their functionality beyond language processing alone. Given this, an intriguing question arises: Can we directly utilize LLMs to improve trajectory prediction without explicit prompt engineering?

To accomplish this objective, we put forward Traj-LLM, a novel trajectory prediction framework, that seeks to investigate the feasibility of LLMs in inferring agents' future trajectories devoid of explicit prompt engineering. We attempt to inject the advanced capabilities of contemporary language models into the trajectory prediction task, providing a general and easily adaptable solution for forecasting in complex driving scenarios based on LLMs. Traj-LLM starts with sparse context joint coding, responsible for parsing the features of agents and scenes into a form that LLMs understand. We then guide LLMs to learn a spectrum of high-level knowledge inherent in trajectory prediction tasks, such as scene context and social interactions. Employing the Parameter-Efficient Fine-Tuning (PEFT) technique, we cost-effectively fine-tune pre-trained LLMs to mitigate the potential disparities between trajectory tokens and natural language texts. Aiming to imitate the human-like lane focus cognitive function and further enhance the scene understanding of Traj-LLM, we present lane-aware probabilistic learning, which is driven by the pioneering Mamba module. Finally, we introduce a multi-modal Laplace decoder to achieve scene-compliant multi-modal predictions. Extensive experiments demonstrate that Traj-LLM, empowered by the robust prior knowledge and inference capabilities of LLMs, along with the human-like lane-aware probability learning, surpasses state-of-the-art methods in evaluation metrics. Furthermore, the few-shot study further validates Traj-LLM's prowess, where even with only half of data volume, it outperforms a majority of baselines reliant on full data utilization. 

The major contributions of our research are summarized below:

\begin{itemize}
\item We propose Traj-LLM, the first trajectory prediction approach powered by pre-trained LLMs without explicit prompt engineering. By integrating the advanced capabilities of LLMs into the trajectory prediction task, we offer a versatile and easily adaptable solution for agent motion forecasting in a new way.     
\item We propose a novel lane-aware probability learning method powered by the pioneering Mamba module, to emulate the humanoid lane-focus cognitive function during the driving decision-making process. This not only enhances the scene understanding skill of Traj-LLM but also explicitly guides motion states to align with potential lane segments.
\item Rigorous experimentation on real-world datasets consistently demonstrates the effectiveness of our model, Traj-LLM. Experimental results indicate that Traj-LLM outperforms the state-of-the-art methods. This performance superiority extends to both overall predictions and the capacity to handle few-shot challenges, affirming the strength of our framework in addressing complex trajectory prediction tasks. 
\end{itemize}

% It is worth mentioning that Mamba formulated as a selective structured state space model, excels in refining and summarizing relevant information via input-dependent adaptations. This mechanism bears resemblance to the sophisticated decision-making process of human drivers, who prudently weigh crucial informational cues, such as lane segments.

\section{Related Work}
\subsection{Trajectory Prediction}
Deep learning-based methodologies have propelled the predictive accuracy of trajectory prediction tasks to unprecedented heights. As a revolutionary study, CS-LSTM \cite{deo2018convolutional} introduces a convolutional social pooling layer to significantly augment vehicle-vehicle interdependencies. In its wake, an abundant number of approaches emerged to capture intricate spatio-temporal interactions \cite{liang2021nettraj,wang2023spatio}. Advanced graph-based proposals, like Graph Convolutional Networks (GCNs) \cite{du2024social,wu2023graph,guo2023vectorized} and Graph Attention Networks (GATs) \cite{zhang2022ai,mo2023map,liao2024cognitive}, harness two pivotal elements, namely, vertexes and edges, to model dynamic spatio-temporal interactions between the target vehicle and its neighbors. Furthermore, Transformer has emerged as a prominent approach for addressing long sequence trajectory prediction problems, thanks to its unique attention mechanism \cite{geng2023physics,hu2023holistic}. Subsequently, in response to data uncertainty and sample multi-modality, a series of research leverages generative models, such as Generative Adversarial Networks (GANs) \cite{roy2019vehicle,li2021vehicle},  Variational Autoencoders (VAEs) \cite{neumeier2021variational,chen2021trajvae} and Conditional Variational Autoencoders (CVAEs) \cite{feng2019vehicle}, to produce multi-modal predictions. 
% CS-LSTM, marking a milestone in the sphere of trajectory prediction, pioneered the incorporation of convolutional social pooling layers to capture the spatio-temporal interactions between the target vehicle and its neighbors \cite{deo2018convolutional}. 

Additionally, a multitude of techniques has been invented to help predict how vehicles will move via lane-based scene information. For example, goal-based models forecast achievable objectives positioned within credible lanes, subsequently constructing entire trajectories \cite{ghoul2022lightweight,yao2023goal}. Alternatively, anchor-based methodologies employ a predetermined collection of anchors aligned with trajectory distribution modes, which facilitates the regression of anticipated multi-modal trajectories \cite{dong2021multi,zhou2023query}. Though these advancements have resulted in impressive performance improvements, a notable gap persists in scene cognition and understanding. The complex driving scene calls for prediction schemes with more powerful modeling to capture a range of scene high-level knowledge and interactive information.

%,li2023conditional,albrecht2021interpretable
%,hasan2021maneuver,kothari2021interpretable,li2023planning

\subsection{Large Language Models}
Recently, LLMs have garnered considerable attention due to their extraordinary comprehension and reasoning prowess in tackling a myriad of tasks, such as prediction \cite{jin2023time,chang2023llm4ts}, classification \cite{gao2024units}, few-shot learning \cite{bian2024multi}, and zero-shot learning \cite{gruver2024large}. Scholars have recognized the powerful reasoning and understanding capabilities natural in LLMs, catalyzing their integration into the sphere of autonomous driving, where they have undergone extensive scrutiny and research. For instance, the integration of vectorized numeric modalities with pre-trained LLMs in LLM-Driver \cite{chen2023driving} substantially promotes the comprehension of intricate traffic scenarios. This showcases the remarkable potential of LLMs in elucidating driving contexts, making informed decisions, and addressing queries adeptly. Human-centric autonomous systems based on LLMs are competent to meet user demands by inferring natural language commands, whereby judicious prompt designs can enhance the efficacy of LLMs \cite{yang2024human}. Drive as You Speak \cite{cui2024drive} perfectly incorporates LLMs into various aspects of autonomous driving, enabling personalized travel experiences and humanoid decision-making, thus propelling autonomous driving towards greater innovation and efficacy.

Meanwhile, a few researchers have initially applied LLMs with prompts to the sphere of trajectory prediction. LG-Traj \cite{chib2024lg} exploits LLMs to guide pedestrian trajectory prediction, integrating motion cues to increase the understanding of pedestrian behavioral dynamics, and adopts a Transformer-based architecture to capture social interactions and learn model representations. LMTraj \cite{bae2024can}, a language-based multi-modal pedestrian trajectory predictor, transforms the trajectory prediction issue into a question-answering task, utilizing LLMs to understand scene contexts and social relationships. These few efforts share a focus on designing pertinent prompts, shifting the paradigm towards prompt-based predictions. However, formulating clear, well-structured, and effective prompts is not a trivial task, which requires consideration of various elements, such as question-answering template \cite{xue2023promptcast}. It potentially introduces modeling bias and integration hurdles. In contrast to these pioneering efforts, we propose to explore the advanced capabilities of LLMs without explicit prompt engineering, thus serving as a more general and easily adaptable solution for trajectory forecasting in complex driving scenarios.

\section{Problem Formulation}

The problem of trajectory prediction is to predict the time-series future coordinates of the target agent over a forthcoming time horizon $t_f$.  Formally, let $\mathcal{X}_i$ denote the $x$ and $y$ positions of the agent $i$ within a designated time horizon $\{-t_h+1,\dots,0\}$ (with $i=0$ for the target vehicle, and $i=1:N$ for surrounding vehicles). The availability of the high-definition (HD) map $\mathcal{M}$, encapsulating scene information, is assumed. Both agents' historical trajectories and lane centerlines are structured as vectorized entities, similar to some established research paradigms \cite{gu2021densetnt,liu2023laformer}. Specifically, for agent $i$, its historical trajectory $\mathcal{X}_i$ is conceptualized as an ordered sequence of sparse trajectory vectors $\mathcal{V}_i=\{v_i^{-t_h+2},\dots,v_i^0\}$ spanning the preceding $t_h$ temporal steps. Each trajectory vector $v_i^t$ comprises the coordinates of the start and end points, denoted as $p_i^{t,s}$ and $p_i^{t,e}$ respectively, alongside attribute features $a_i$ (e.g., object type and timestamps). Furthermore, to ensure input feature invariance regarding an agent's location, the coordinates of all vectors are normalized to be centered around the target agent's most recent position. For capturing sophisticated lane information, lane centerlines abstracted by polylines are divided into predefined segments. In this way, the lanes contain a variable number of vectors denoted as $\mathcal{M}_i^{1:L}=\{v_i^1,\dots,v_i^L\}$, where $L$ signifies the total vector length. The lane vector $v_i^l$ is annotated with sampled points ($p_i^{l,s}$ and $p_i^{l,e}$), attribute attributes $a_i$, and indicator $p_i^{l,pre}$ standing for the predecessor of the start point.

Given the map and agent states, our goal is tasked with forecasting the trajectory conditional distribution $P(\mathcal{Y}|\mathcal{X},\mathcal{M})$ for the subsequent $t_f$ time intervals, where $\mathcal{Y} = \left\{ y_0^{1}, y_0^{2}, \dots, y_0^{t_f} \in \mathbb{R}^{t_f \times 2} \right\}$. It is postulated that the distribution of $y_0^{t_f}$ conforms to a Laplace distribution. In this work, we are dedicated to generating $K$ future trajectories for each target agent and apportioning a probability score for each prediction.
% For perspicuity, Table 1 delineates the principal variables and parameters expounded upon in this paper.

\begin{figure*}
  \centering
    \includegraphics[width=7.1in]{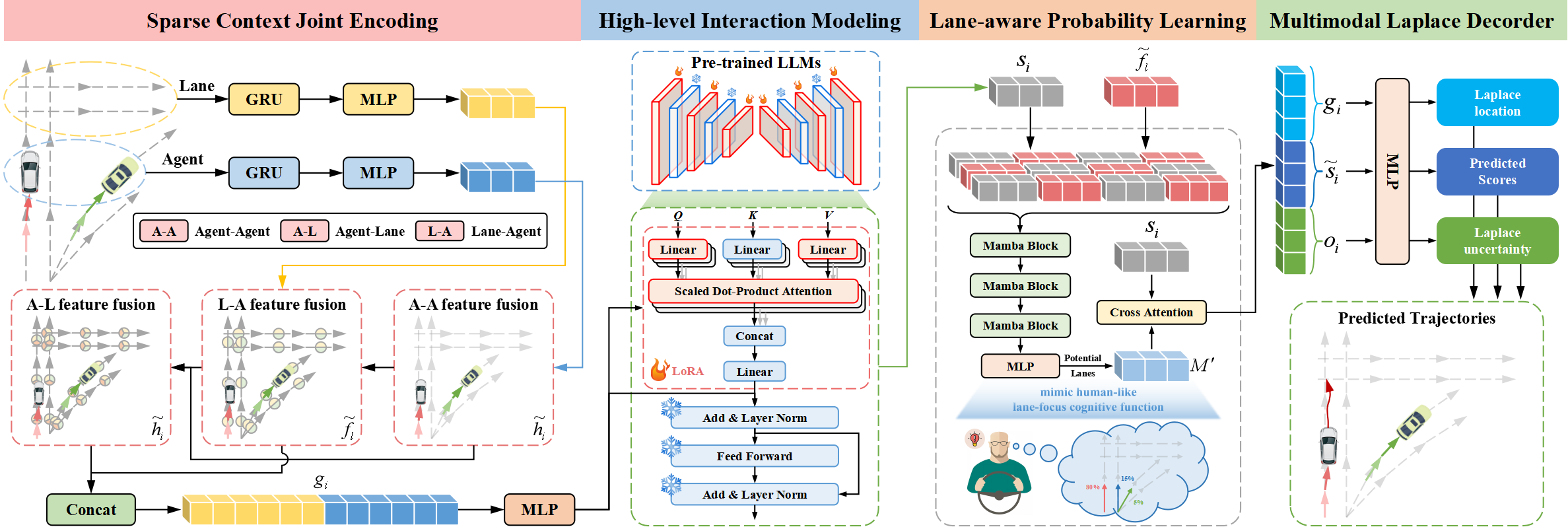}
    \caption{Framework of Traj-LLM.}
    \label{fig: framework}
\end{figure*}

% \begin{table}[htpb]
% \renewcommand\arraystretch{1}
% \centering
% \caption{\centering Parameter settings.}
% \setlength{\tabcolsep}{0.6mm}
% \footnotesize
% \begin{tabular}{ll}
% \toprule
% \textbf{Parameter} & \textbf{Description} \\
% \midrule
% $t_f$ & The given future time horizon\\
% $t_h$ & The given history time horizon\\
% $\mathcal{X}_i$ & $x$ and $y$ positions of the agent $i$ in a driving scenario \\
% $\mathcal{M}$ & The HD map\\
% $\mathcal{V}_i$ & The sequence of sparse trajectory vectors for the agent $i$\\
% $v_i^t$ & Each trajectory vector\\
% $p_i^{t,s}$ & The coordinates of the start points\\
% $p_i^{t,e}$ & The coordinates of the endpoints\\
% $a_i$ & The attribute features\\
% $v_i^l$ & The lane vector\\
% $p_i^{l,pre}$ & The indicator signifying the predecessor of the start point\\
% $\mathcal{Y}$ & The predicted future trajectory of the target agent\\
% $K$ & The number of predicted trajectories \\

% \bottomrule
% \end{tabular}
% \label{tab:parameter}
% \end{table}

\section{Proposed Model}

The overarching architecture of Traj-LLM is illustrated in Fig. \ref{fig: framework}, which contains four integral components: sparse context joint encoding, high-level interaction modeling, lane-aware probability learning, and multi-modal Laplace decoder. Our approach stands as a trailblazer in exploring the ability of LLMs for trajectory prediction tasks dispensing with the need for explicit prompt engineering. The sparse context joint coding initially engages in parsing the features of agents and scenes into a form understandable by LLMs. Subsequently, the resulting representations are fed into pre-trained LLMs to address high-level interactions. To mimic the human-like lane focus cognitive function and further enhance the scene understanding of Traj-LLM, lane-aware probability learning is presented based on the well-designed Mamba module. Finally, a multi-modal Laplace decoder is used to generate reliable predictions. Below, we describe each module of Traj-LLM in detail.

\subsection{Sparse Context Joint Encoding}
The first step of Traj-LLM is to encode the spatial-temporal scene input, such as agent states and lanes. For each of them, we employ an embedding network that consists of a GRU layer and an MLP to extract the high-dimensional features. Thereafter, the resulting tensors $h_i$ and $f_l$ are channeled into the fusion submodule, facilitating the complex information exchange between the agent states and lanes within localized regions. This process is performed in a token embedding-like manner to align with the way LLMs work.

More specifically, the fusion process entails the utilization of the multi-head self-attention mechanism ($MultiSelAtt$) for agent-agent feature fusion, followed by Gated Linear Units \cite{dauphin2017language} ($GLU$). Additionally, the fusion of lane-agent and agent-lane features involves updating the agent and lane representations through the multi-head cross-attention mechanism ($MultiCrossAtt$) with skip connections. This process can be formally expressed as follows:
\begin{equation}
\widetilde{h_i}=MultiSelfAtt(h_i,h_i),i \in \{0,\dots,N\},
\end{equation}
\begin{equation}
\widetilde{h_i}=GLU(h_i,\widetilde{h_i}),i \in \{0,\dots,N\},
\end{equation}
\begin{equation}
\widetilde{f_l}=f_l+MultiCrossAtt(f_l,\widetilde{h_i}),l \in \{0,\dots,L\},
\end{equation}
\begin{equation}
\widetilde{h_i}=\widetilde{h_i}+MultiCrossAtt(\widetilde{h_i},\widetilde{f_l}),i \in \{0,\dots,N\}.
\end{equation}

Eventually, we concate $\widetilde{h_i}$ and $\widetilde{f_l}$ to produce sparse context joint encodings $g_i$, which intuitively carry dependencies pertinent to local receptive fields among the vectorized entities. The sparse context joint encoding is designed to enable LLMs to understand trajectory data, thereby boosting the advanced capabilities of LLMs.

\subsection{High-level Interaction Modeling}
Trajectory transitions adhere to patterns governed by high-level constraints emanating from various elements in the scene. To learn these high-level interactions, we explore the capacity of LLMs to model a range of dependencies inherent in trajectory prediction tasks. Despite the semblance between trajectory data and natural language texts, direct utilization of LLMs to handle sparse context joint encodings is deemed unsuitable, given that generic pre-trained LLMs are primarily tailored for textual data processing. One alternative proposal is to undergo comprehensive retraining of the entire LLMs, a process demanding substantial computational resources, thus rendering it somewhat impracticable. Another more efficient solution lies in the application of the Parameter-Efficient Fine-Tuning (PEFT) technique to fine-tune pre-trained LLMs. By adjusting or introducing trainable parameters, PEFT showcases the outstanding capability to optimize pre-trained LLMs for downstream tasks by a large margin \cite{zhang2023llama,zhang2022tip}.

\begin{figure*}
  \centering
    \includegraphics[width=5.3in]{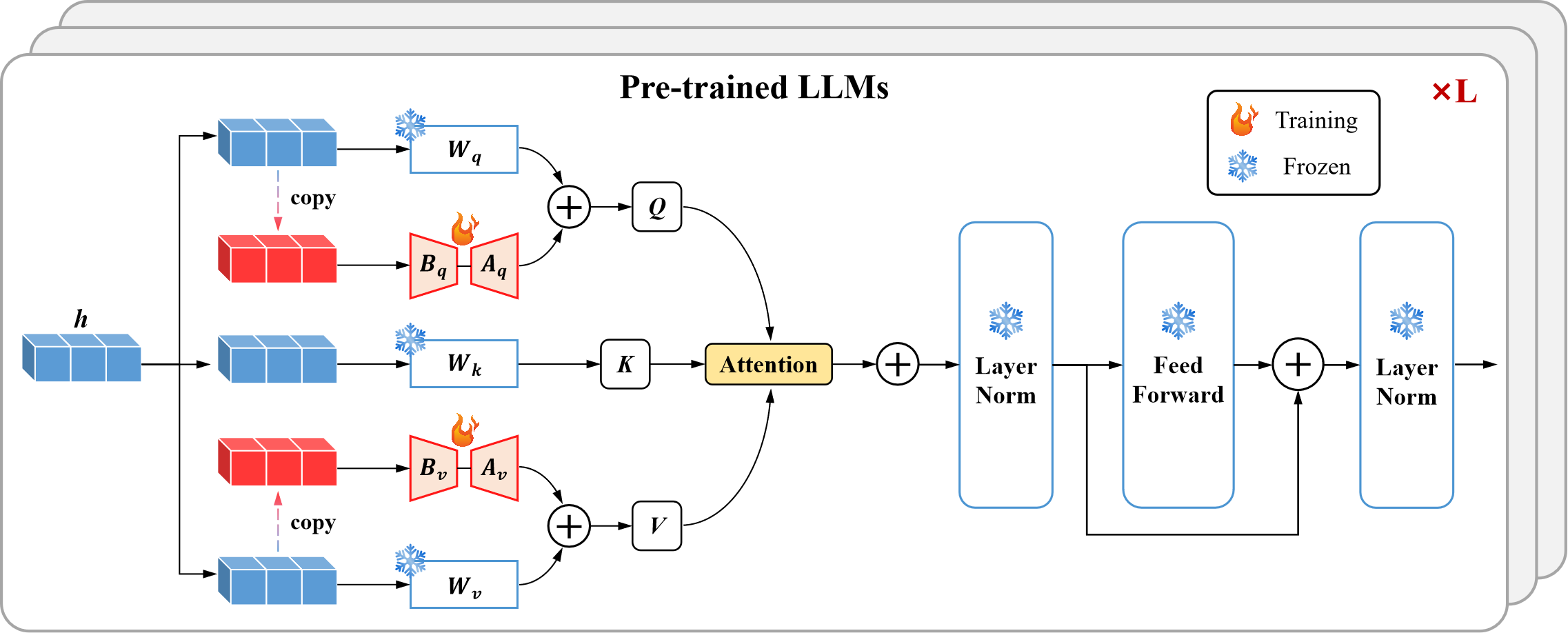}
    \caption{The overview of Pre-trained LLMs.}
    \label{fig: LLM}
\end{figure*}

In this research, we utilize parameters from NLP pre-trained transformer architectures, specifically focusing on the GPT2 model \cite{radford2019language} as shown in Fig. \ref{fig: LLM}, for high-level interaction modeling. We opt to freeze all pre-trained parameters and inject new trainable ones by implementing the Low-Rank Adaptation (LoRA) technique \cite{hu2021lora}. LoRA is applied to the Query and Key of the attention layers in LLMs. Let denote the rank of LoRA as $r$, the input of Query and Key in $j$-$th$ attention layer $\theta_j$ as $a_j$ with size $d$, and the output as $\widetilde{a_j}$ with size $k$. For a pre-trained weight matrix $W \in \mathbb{R}^{d \times k}$ in the network $\theta_j$, LoRA approximately updates it using:
\begin{equation}
\widetilde{W} \approx W + BA,
\end{equation}
where the rank decomposition matrix $B \in \mathbb{R}^{d \times r}$ and $A \in \mathbb{R}^{r \times k}$, with the rank $r\ll min(d,k)$. During the training process, $W \in \mathbb{R}^{d \times k}$ is kept frozen, while $B$ and $A$ are treated as trainable parameters,  initialized to zero and Gaussian distributions, respectively. Therefore, the forward transfer function of LoRA can be succinctly expressed as:
\begin{equation}
\widetilde{a_j} = W\cdot{a_j} + BA\cdot{a_j}.
\end{equation}

On this basis, we pass the given sparse context joint encodings $g_i$ into pre-trained LLMs, which own a series of pre-trained transformer blocks equipped with LoRA. This procedure ends up with the generation of high-level interaction representations $z_i$: 
\begin{equation}
z_i=LLMs(g_i).
\end{equation}

After being processed by pre-trained LLMs, the output representations $z_i$ are transformed via an MLP layer to match the dimensions of $g_i$, thus yielding the ultimate high-level interaction states $s_i$.

\begin{figure}
  \centering
    \includegraphics[width=3.3in]{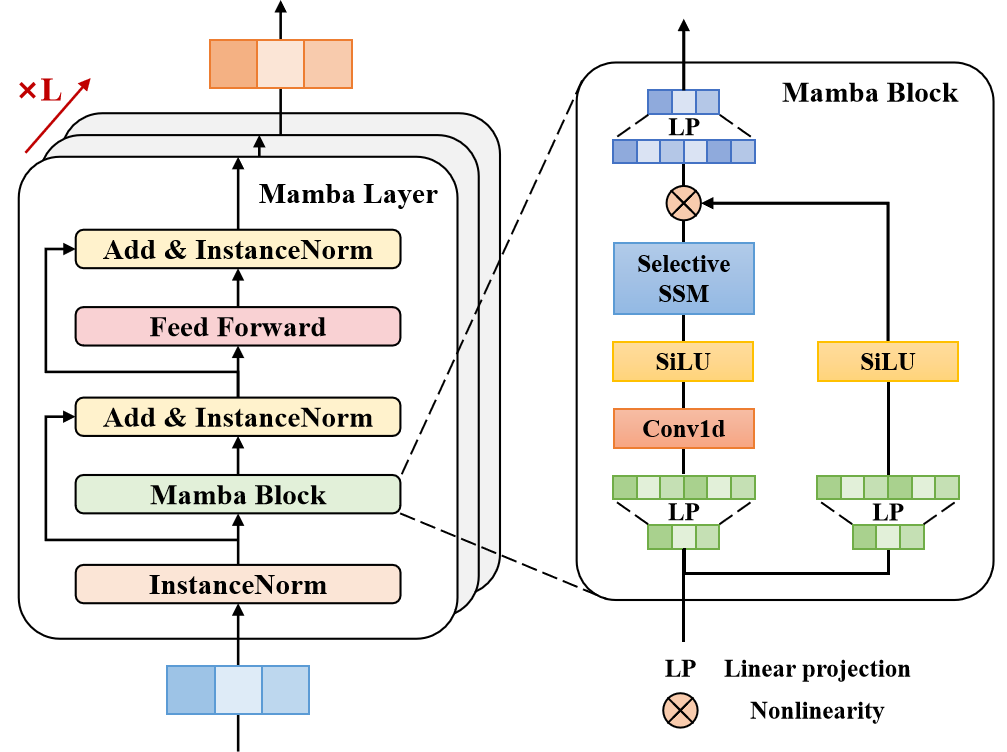}
    \caption{The proposed Mamba layer for lane-aware probability learning.}
    \label{fig: mamba}
\end{figure}

\subsection{Lane-aware Probability Learning }
The overwhelming majority of experienced drivers merely pay attention to a few potential lane segments, which exert remarkable influence on their future motions. To mimic this human-like cognitive function and further enhance the scene understanding of Traj-LLM, we employ lane-aware probabilistic learning to continuously estimate the likelihood of motion states aligning with lane segments as shown in Algorithm 1. Precisely, we synchronize the target agent's motion with lane information at each time step $t \in \{1,\dots,t_f\}$ with introducing the Mamba layer. Functioning as a selective structured state space model (SSM), Mamba excels at refining and summarizing pertinent information rather than indiscriminately traversing all sequences \cite{gu2023mamba}. It is analogous to the sophisticated decision-making of human drivers, who judiciously weigh key informational cues, such as potential lanes, to inform their driving choices. 

Fig. \ref{fig: mamba} depicts the Mamba layer, which consists of a Mamba block, three-layer normalization, and a position-wise feed-forward network. The Mamba layer first normalizes the input $F=concat[s_i,\widetilde{f_l}]\in \mathbb{R}^{B \times L \times D}$, whose dimensions are determined by the batch size $B$, the length of lane segments $L$, and the hidden dimension $D$. Then the Mamba block maps the normalized $F\in \mathbb{R}^{B \times L \times D}$, to the output $Q\in \mathbb{R}^{B \times L \times D}$:
\begin{equation}
Q = Mamba(InstanceNorm(F)).
\end{equation}

Concretely, as granularly detailed in Algorithm 1, the Mamba block first expands the input dimension with dilatation coefficient $E$ through linear projections, yielding distinct representations for two parallel processing branches, designated as $m$ and $n$. One branch is then processed through a 1D convolution and a SiLU activation \cite{elfwing2018sigmoid}, to capture lane-aware dependencies $m'$. The core of the Mamba block involves a selective state space model with parameters discretized based on the input. The $m'$ is linearly projected to the $B$, $C$, $\Delta$, respectively. The $\Delta$ is then used to transform the $\overline{A},\overline{B}$. And SelectiveSSM accepts $m'$ and $\overline{A},\overline{B},C$ as input, generating refined lane-aware features $q$. Simultaneously, the other branch introduces a simple SiLU activation to produce gating signals $n'$, intended for filtering extraneous information. Finally, $q$ is multiplicated with $n'$, followed by a linear projection to deliver the ultimate output $Q$.

% %The SSM maps the input $F$ to output variates $F'$ through intermediate implicit states $h$:
% \begin{equation}
% \begin{aligned}
% h' &= Ah+BF, \\
% F' &= Ch,
% \end{aligned}
% \end{equation}
% where $A,B,C$ are learnable matrics. The SSM further discretizes the continuous sequence using a step size $\Delta$:
% \begin{equation}
% \begin{aligned}
% h' &= \overline{A}h+\overline{B}F, \\
% F' &= Ch,
% \end{aligned}
% \end{equation}
% where $\overline{A}=\text{exp}(\Delta{A})$ and $\overline{B}=(\Delta{A})^{-1}(\text{exp}(\Delta{A})-i)\cdot\Delta{B}$. 

To enhance robustness, we further incorporate the instance normalization and residual connection to get implicit states $\widetilde{Q}$:
\begin{equation}
\widetilde{Q} = InstanceNorm(Dropout(Q))+F.
\end{equation}

Subsequently, we leverage a position-wise feed-forward network to improve the modeling of lane-aware estimation in the hidden dimension:
\begin{equation}
\widetilde{Q'} = ReLU(\widetilde{Q}W^{(0)}+b^{(0)})W^{(1)}+b^{(1)},
\end{equation}
where $W^{(0)},W^{(1)},b^{(0)},b^{(1)}$ are trainable parameters. Once again, the instance normalization and residual connection are executed to acquire the lane-aware learning vectors $S$:
\begin{equation}
S = InstanceNorm(Dropout(\widetilde{Q'}))+\widetilde{Q},
\end{equation}
which are then fed into an MLP layer, resulting in the predicted score of the $l$-th lane segment at $t$:
\begin{equation}
    p_{l,t} = \frac{\exp(MLP(S_l))}{\sum_{j=1}^{L} \exp(MLP(S_j))}.
\end{equation}

As elaborated above, skilled drivers exhibit a keen focus on multiple potential lane segments to facilitate effective decision-making. To this end, we carefully curate the top-c lane segments $\{v_1,v_2,\dots,v_c\}$ with the top-$c$ scores $\{p_1,p_2,\dots,p_c\}$ as the candidate lane segments, which are further concatenated to form $\mathcal{M}'$. 

The lane-aware probability learning is modeled as a classification problem, wherein a binary cross-entropy loss $\mathcal{L}_{lane}$ is applied to optimize the probability estimation:
\begin{equation}
\mathcal{L}_{lane} = \sum_{t=1}^{t_f}\mathcal{L}_{CE}(p_t,\widetilde{p_t}).
\end{equation}

Here, the ground truth score $\widetilde{p_t}$ is set to 1 for the lane segment closest to the trajectory's truth position, while 0 for others. The distance $d(l,y_{0})$ between a lane segment $l$ and the ground truth position is discerned through Euclidean distance calculation:
\begin{equation}
d(l,y_{0}) = || l-y_0||^2.
\end{equation}

\begin{algorithm}
  \caption{Lane-aware probability learning}
 \KwIn{$F:(B,L,D)$}
 \KwOut{$\mathcal{M}':\{v_1,v_2,\dots,v_c\}$}
  $m:(B,L,ED)\leftarrow{\text{Linear}^m(F)\{\text{Linear projection}\}}$\;
  $n:(B,L,ED)\leftarrow{\text{Linear}^n(F)\{\text{Linear projection}\}}$\;
  $m':(B,L,ED)\leftarrow{\text{SiLU}(\text{Conv1d}(m))}$\;
  $A:(D,D')\leftarrow{\text{Parameter}\{\text{Structured state matrix}\}}$\;
  $B:(B,L,D')\leftarrow{\text{Linear}^B(m')\{\text{Linear projection}\}}$\;
  $C:(B,L,D')\leftarrow{\text{Linear}^C(m')\{\text{Linear projection}\}}$\;
  $\Delta:(B,L,D)\leftarrow{\text{Softplus}(\text{Parameter}+\text{Broadcast}(\text{Linear}^{\Delta}(m')))}$\;
  $\overline{A},\overline{B}:(B,L,D,D')\leftarrow{\text{discretize}(\Delta,A,B)}$\;  
  $q:(B,L,ED)\leftarrow{\text{SelectiveSSM}(\overline{A},\overline{B},C)(m')}$\; %(\overline{A},\overline{B},C)
  $n':(B,L,ED)\leftarrow{\text{SiLU}(n)}$\;
  $Q:(B,L,D)\leftarrow{\text{Linear}(q\otimes{n'})\{\text{Linear projection}\}}$\;
  $\widetilde{Q}:(B,L,D)\leftarrow{\text{InstanceNorm}(\text{Dropout}(Q))+F}$\;
  $\widetilde{Q'}:(B,L,D)\leftarrow{\text{ReLU}(\widetilde{Q}W^{(0)}+b^{(0)})W^{(1)}+b^{(1)}}$\;
  $S:(B,L,D)\leftarrow{\text{InstanceNorm}(\text{Dropout}(\widetilde{Q'}))+\widetilde{Q}}$\;
  $p_{l,t}:(B,L)\leftarrow{ \frac{\exp(MLP(S_l))}{\sum_{j=1}^{L} \exp(MLP(S_j))}}$\;
  $\mathcal{M}'\leftarrow{\text{choose} \{v_1,v_2,\dots,v_c\}}$\;
  \textbf{return} $\mathcal{M}'$
\end{algorithm}

\subsection{Multi-modal Laplace Decoder}
The anticipated movements of traffic agents are inherently multi-modal. Hence, we adopt a mixture model framework to parameterize the prediction distribution, where each mixture component follows a Laplace distribution, in line with established methodologies \cite{zhou2022hivt,zhou2023query}. For each predicted instance, the multi-modal Laplace decoder takes the representations $e_i$ as inputs and outputs a set of trajectories $\sum_{k=1}^{K}{\pi}_{i,k}\prod \limits_{t=1}^{t_f} \text{Laplace}(\mu_i^t,b_i^t)$. Here, $\{{\pi}_{i,k}\}_{k=1}^{K}$ are the mixing coefficients, and the $k$-th mixture component's Laplace density is parameterized by the location $\mu_i^t \in \mathbb{R}^2$ and its associated uncertainty $b_i^t \in \mathbb{R}^2$. The representations $e_i$ are made of the sparse context joint encodings $g_i$, the lane-aware guided high-level interaction feature $\widetilde{s_i}$, together with a latent vector $o$ sampled by a multivariate normal distribution. The feature $\widetilde{s_i}$ is generated via cross-attention between $s_i$ and $\mathcal{M}'$, to guide the target agent towards candidate lane segments like a skilled driver. To predict mixing coefficients, an MLP followed by a softmax function is employed, while two side-by-side MLPs are utilized to generate $\mu_i^t$ and $b_i^t$. Then, we use a regression loss and a classification loss to train the multi-modal Laplacian decoder. The regression loss is computed using the Winner-Takes-All strategy \cite{zhou2022hivt,liu2023laformer}, defined as: 
\begin{equation}
\mathcal{L}_{reg} = -\frac{1}{t_f}\sum_{t=1}^{t_f}\log{P}\left(y_{0}^{t} | \mu^{t}_{i,k^*}, b^{t}_{i,k^*}\right)
\end{equation}
where $y_0^t$ represents the ground truth position, and $k^{*}$ signifies the mode with the minimum $L_2$ error among the $K$ predictions. On the other hand, the cross-entropy loss is adapted as the classification loss $\mathcal{L}_{cls}$ to adjust the mixing coefficients.  

Consequently, the overall loss function $\mathcal{L}$ of Traj-LLM can be written as:
\begin{equation}
\mathcal{L} = \lambda\mathcal{L}_{lane}+\mathcal{L}_{reg}+\mathcal{L}_{cls},
\end{equation}
where $\lambda$ serves as a hyper-parameter, controlling the relative importance of $\mathcal{L}_{lane}$.

\section{Experiment}
This section delineates a comprehensive series of experiments aimed at substantiating the efficacy of Traj-LLM. We commence by elucidating the experimental setup, followed by a meticulous juxtaposition of results compared with various cutting-edge approaches. To ascertain the impact of each constituent within our model, we conduct ablation experiments on Traj-LLM architectures. Furthermore, we embark on the few-short study to show the generalization capability of Traj-LLM. Lastly, we delve into a qualitative analysis of predictions.

\subsection{Experimental Setup}

\textbf{Evaluation Datasets.} The assessment of Traj-LLM is conducted on a widely used benchmark, the nuScenes dataset \cite{caesar2020nuscenes}, for trajectory prediction. This dataset covers 1,000 driving scenarios spanning different cities (e.g., Boston and Singapore), which were collected via vehicle-mounted cameras and lidar sensors operating at a sampling frequency of 2 Hz, providing an exhaustive depiction of urban traffic dynamics. Adhering to the official predictions scheme, Traj-LLM adopts 2-second segments of sequences to forecast subsequent 6-second trajectories. Furthermore, the predicted modality $K$ is set to 5 and 10, which also aligns harmoniously with the specifications and recommendations of the nuScenes dataset.

\textbf{Evaluation Metrics.} We subject our model to rigorous evaluation using standard metrics for motion prediction, including the minimum Average Displacement Error (minADE), minimum Final Displacement Error (minFDE), and Miss Rate (MR). The minADE metric quantifies the $\ell_2$ distance in meters between the best-predicted trajectory and its corresponding ground-truth trajectory, averaged across all points. At the same time, minFDE measures the discrepancy at the final position. MR denotes the proportion of instances where the distance between the actual endpoint and the optimal predicted endpoint exceeds 2.0 meters. In all cases, diminished values signify heightened efficacy in model performance. 

\textbf{Implementation Details.} Our model is trained on 6 NVIDIA GeForce RTX4090 GPUs with the AdamW optimizer, whose batch size and initial learning rate are set to 132 and 0.001, respectively. The architecture of Traj-LLM is carefully designed, beginning with a layer of sparse context joint encoding module, followed by a high-level interaction modeling module, and ultimately incorporating three layers of lane-aware probability learning module. The hidden dimensions of all feature vectors are uniformly configured to 128.  
% The rank of LoRA is 64 and the hyper-parameter $\lambda$ is 0.9.
\begin{table}[]
\renewcommand\arraystretch{1.3}
\caption{\centering The comparison result of the proposed Traj-LLM and state-of-art methods for $K=5$. The best/second best values are highlighted in boldface/underlined.}
\setlength{\tabcolsep}{5.1mm}
\begin{tabular}{@{}l|ccc@{}}
\toprule
\multicolumn{1}{@{}l|}{Method} & $\text{minADE}_5$& $\text{minFDE}_5$& $\text{MR}_5$\\ \midrule
Trajectron++ \cite{salzmann2020trajectron++}&                  1.88 &                  -&                  0.70 \\
ALAN \cite{narayanan2021divide}&                               1.87 &              3.54 &                  0.60 \\
GATraj \cite{cheng2023gatraj}&                                 1.87 &              4.08 &                   -   \\
SG-Net \cite{wang2022stepwise}&                                1.85 &              3.87 &                   -   \\
WIMP \cite{khandelwal2020if} &                                 1.84 &               -   &                  0.55 \\
MHA-JAM \cite{messaoud2021trajectory}&                         1.81 &              3.72 &                  0.59 \\
AgentFormer \cite{yuan2021agentformer}&                        1.59 &              3.14 &                    -  \\
LaPred \cite{kim2021lapred}&                                   1.47 &               -   &                  0.53 \\
P2T \cite{deo2020trajectory}&                                  1.45 &               -   &                  0.64 \\
GOHOME \cite{gilles2022gohome}&                                1.42 &               -   &                  0.57 \\
CASPNet \cite{schafer2022context}&                             1.41 &               -   &                  0.60 \\
MUSE-VAE \cite{lee2022muse}&                                   1.38 &   \underline{2.90}&                    -  \\
Autobot \cite{girgis2021latent}&                               1.37 &               -   &                  0.62 \\
THOMAS \cite{gilles2021thomas}&                                1.33 &               -   &                  0.55 \\
HLSTrajForecast \cite{choi2022hierarchical}&                   1.33 &              2.92 &                    -  \\
PGP \cite{deo2022multimodal}&                                  1.27 &               -   &                  0.52 \\
LAformer \cite{liu2023laformer} &                      \textbf{1.19}&               -   &       \underline{0.48}\\ \midrule
Traj-LLM&                                           \underline{1.24}&      \textbf{2.46}&          \textbf{0.41}\\ %\midrule
% \toprule[1pt]
\bottomrule
\end{tabular}
\label{tab: nuscene5}
\end{table}

\subsection{Results and Computational Performance}

\textbf{Comparison with State-of-the-art.} We benchmarked Traj-LLM against a plethora of the latest state-of-the-art models, involving various architectural paradigms. The quantitative results are illustrated in Table \ref{tab: nuscene5} for $K=5$ and \ref{tab: nuscene10} for $K=10$. It is discernible that Traj-LLM displays excellent performance, essentially securing the top or second position across all metrics, thanks to the advanced capabilities of LLMs and well-designed lane-aware probability learning. In particular, when scrutinizing the metrics $\text{minFDE}_5$ and $\text{MR}_5$, Traj-LLM demonstrates substantial enhancements over MUSE-VAE and LAformer, elevating performance by 15.17\% and 14.58\%, respectively. Similarly, Traj-LLM exhibits notable enhancements in predictive accuracy on the metrics $\text{minFDE}_{10}$ and $\text{MR}_{10}$, outperforming the suboptimal approaches, ALAN and LAformer, by substantial margins of 7.49\% and 30.30\%, respectively. Furthermore, GOHOME stands out as the rasterized method in the table, yet it notably lags behind most vectorized counterparts in performance metrics. Compared with other lane-based methodologies like LaPred, Traj-LLM maintains a clear superiority, underscoring the efficacy of our lane-aware probability learning in mimicking human-like lane-focus cognitive function and enhancing scene understanding. It is worth highlighting that in contrast to models based on GCN to extract interaction dependencies, such as GATraj, Traj-LLM reduces the $\text{minADE}_5/\text{minFDE}_5\ (\text{minADE}_{10}/\text{minFDE}_{10})$ from 1.87/4.08 (1.46/2.97) to 1.24/2.46 (0.99/1.73). This indicates the pivotal role of pre-trained LLMs in facilitating nuanced feature comprehension and augmenting predictive accuracy. The findings advocate for the integration of LLMs into trajectory prediction tasks, an emerging domain for further exploration. 

\begin{table}[]
\renewcommand\arraystretch{1.3}
\caption{\centering The comparison result of the proposed Traj-LLM and state-of-art methods for $K=10$. The best/second best values are highlighted in boldface/underlined.}
\setlength{\tabcolsep}{4.5mm}
\begin{tabular}{@{}l|ccc@{}}
\toprule
\multicolumn{1}{@{}l|}{Method} & $\text{minADE}_{10}$& $\text{minFDE}_{10}$& $\text{MR}_{10}$\\ \midrule
Trajectron++ \cite{salzmann2020trajectron++}&                  1.51 &                     -  &                  0.57 \\
GATraj \cite{cheng2023gatraj}&                                 1.46 &                   2.97 &                    -  \\
SG-Net \cite{wang2022stepwise}&                                1.32 &                   2.50 &                    -  \\
AgentFormer \cite{yuan2021agentformer}&                        1.31 &                   2.48 &                    -  \\
MHA-JAM \cite{messaoud2021trajectory}&                         1.24 &                   2.21 &                  0.45 \\
ALAN \cite{narayanan2021divide}&                               1.22 &        \underline{1.87}&                  0.49 \\
CASPNet \cite{schafer2022context}&                             1.19 &                     -  &                  0.43 \\
P2T \cite{deo2020trajectory}&                                  1.16 &                     -  &                  0.46 \\
GOHOME \cite{gilles2022gohome}&                                1.15 &                     -  &                  0.47 \\
LaPred \cite{kim2021lapred}&                                   1.12 &                     -  &                  0.46 \\
WIMP \cite{khandelwal2020if}&                                  1.11 &                     -  &                  0.43 \\
MUSE-VAE \cite{lee2022muse}&                                   1.09 &                   2.10 &                    -  \\
THOMAS \cite{gilles2021thomas}&                                1.04 &                     -  &                  0.42 \\
HLSTrajForecast \cite{choi2022hierarchical}&\multicolumn{1}{c}{1.04}&\multicolumn{1}{c}{2.15}& \multicolumn{1}{c}{-} \\
Autobot \cite{girgis2021latent}&                               1.03 &                     -  &                  0.44 \\
PGP \cite{deo2022multimodal}&                       \underline{0.94}&                     -  &                  0.34 \\
LAformer \cite{liu2023laformer}&                       \textbf{0.93}&                     -  &       \underline{0.33}\\ \midrule
Traj-LLM&                                                      0.99 &           \textbf{1.73}&          \textbf{0.23}\\ %\midrule
% \toprule[1pt]
\bottomrule
\end{tabular}
\label{tab: nuscene10}
\end{table}

\begin{figure}
  \centering
    \includegraphics[width=3.3in]{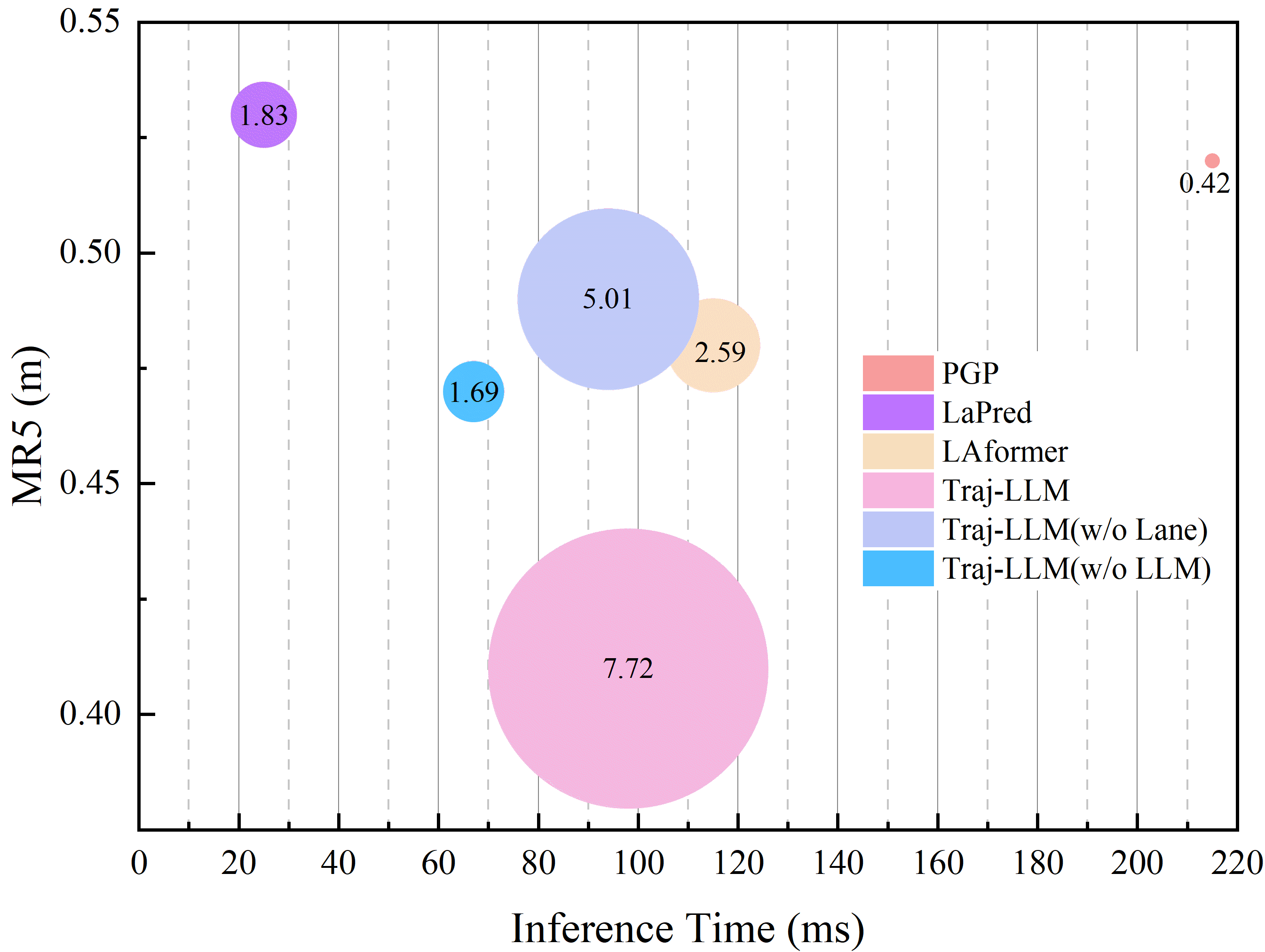}
    \caption{Comparison of Traj-LLM with baseline models across three key metrics: trainable parameters, inference speed, and $\text{MR}_5$. The size of the circles in the figure corresponds to the number of trainable parameters in each model.}
    \label{fig: inference speed mr5}
\end{figure}

\begin{figure}
  \centering
    \includegraphics[width=3.3in]{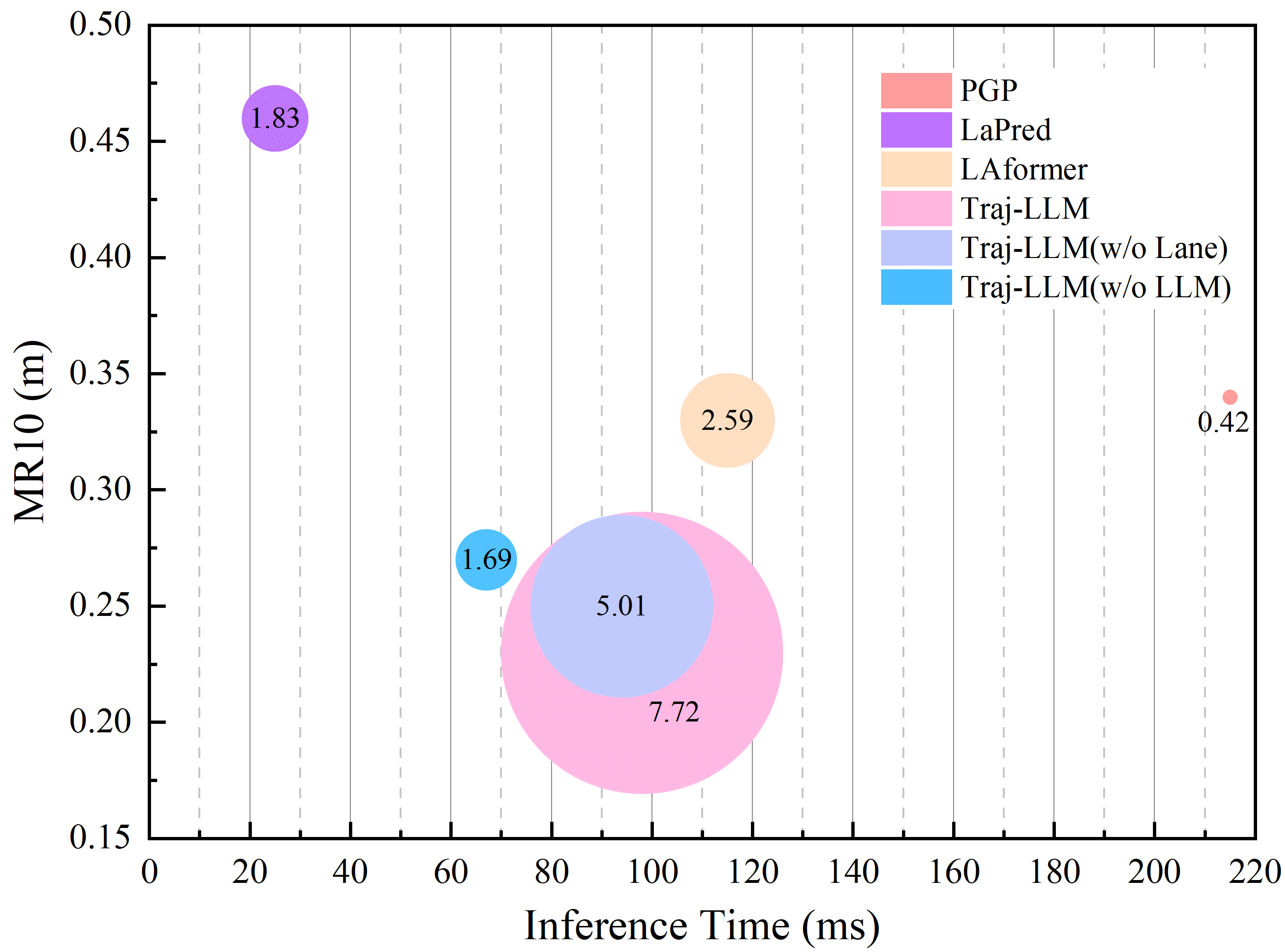}
    \caption{Comparison of Traj-LLM with baseline models across three key metrics: trainable parameters, inference speed, and $\text{MR}_{10}$. The size of the circles in the figure corresponds to the number of trainable parameters in each model.}
    \label{fig: inference speed mr10}
\end{figure}

\textbf{Computational Performance.} To ascertain the efficiency of Traj-LLM, we carried out a comparative analysis of the trainable parameters and inference speed, as shown in Table \ref{tab: parameter}. Based on the pre-trained Large Language Model, the full Traj-LLM owns 7.72M trainable parameters, a magnitude surpassing that of its counterparts. However, Traj-LLM exhibits a pronounced advantage in inference time over LAformer and PGP, despite their comparatively diminutive parameter count. Remarkably, Traj-LLM achieves an impressive inference time of 98 ms, in a scenario featuring an average of 12 agents. The variants of Traj-LLM, including Traj-LLM (w/o Lane) and Traj-LLM (w/o LLM), with trainable parameters of 5.01M and 1.69M, respectively, boast parameter scales either on par with or smaller than those of the alternative models. Yet they also maintain a superior edge in inference speed. Fig. \ref{fig: inference speed mr5} and Fig. \ref{fig: inference speed mr10} further emphasize the prominent equilibrium between prediction accuracy and inference efficiency of our methodology.

\begin{table}[htpb]
\renewcommand\arraystretch{1.2}
    \centering
    \caption{\centering The result of the computational performance. }
    \setlength{\tabcolsep}{2.2mm}{
    \begin{tabular}{@{}l|ccc@{}}
    \toprule
    \multicolumn{1}{@{}l|}{Model} & Trainable params & Batch size &   Inference Speed   \\ \midrule
    PGP \cite{deo2022multimodal}    &     0.42M      &    12      &        215ms        \\
    % P2T \cite{deo2020trajectory}  &       -        &     4      &        79ms         \\ 
    LaPred \cite{kim2021lapred}     &     1.83M      &    16      &        25ms         \\
    % AgentFormer                   &     0.58M      &    64      &        133ms        \\
    LAformer \cite{liu2023laformer} &     2.59M      &    12      &        115ms        \\   
    $\text{Traj-LLM(w/o Lane)}^1$   &     5.01M      &    12      &        94ms         \\    
    $\text{Traj-LLM(w/o LLM)}^2$    &     1.69M      &    12      &        67ms         \\
    Traj-LLM                        &     7.72M      &    12      &        98ms         \\
    \bottomrule
    \end{tabular}}

    \begin{tablenotes}
    \footnotesize
    \item $^1$ Traj-LLM(w/o Lane): the lane-aware probability learning is removed. \\
    \item $^2$ Traj-LLM(w/o LLM): the high-level interaction module is removed.
    \end{tablenotes}
    \label{tab: parameter}
\end{table}

\subsection{Ablation Studies}

We undertook thorough ablation experiments to gauge the contribution of each component in our proposed method. These experiments entailed designing and evaluating several variants of Traj-LLM to check the impact of each modification on the network's overarching performance: 

\begin{itemize}
\item [1)] {w/o LLM: This variant directly removes the entire high-level interaction module, allowing the model to make predictions independent of LLMs.}

\item [2)] {w/o LoRA: This variant retains the high-level interaction module but resorts to freezing LLMs for inferring agent motions, intended to explore the function of the Parameter-Efficient Fine-Tuning (PEFT) technique.}

\item [3)] {w/o Lane: This variant eliminates the lane-aware learning module, thus ignoring the humanoid decision-making process of selecting multiple candidate lanes to guide motions.}

\end{itemize}

The ablation results, shown in Table \ref{tab: ablation result nuscene5} and Table \ref{tab: ablation result nuscene10}, yield discernible insights. Firstly, the variant devoid of LLMs exhibits a weaker performance, manifesting a significant 7.52\% and 9.42\% reduction in accuracy for $\text{minFDE}_5$ and $\text{minFDE}_{10}$, separately, under both $K$=5 and $K$=10 conditions. This confirms the potential capability of LLMs in boosting trajectory prediction tasks, even absent explicit design prompt engineering. Secondly, it becomes evident that fine-tuning LLMs is necessary to effectively transfer cross-modal inference capabilities to comprehension of complex traffic scenes. Last but not least, the humanoid lane-selection mechanism tremendously elevates goal-directed agent navigation accuracy, as reflected by the metric of $\text{MR}$. Specifically, the variant without the lane-aware learning module displays a remarkable 12.77\% and 14.81\% accuracy decrease in $\text{MR}_5$ and $\text{MR}_{10}$, respectively, compared to Traj-LLM. The experimental results corroborate the indispensable roles of both LLMs and human-like lane-aware learning mechanisms in improving model performance.

\begin{table}[htpb]
\renewcommand\arraystretch{1.2}
    \centering
    \caption{\centering The result of the Traj-LLM and its variants in the ablation experiment for $K$=5.}
    \setlength{\tabcolsep}{3.5mm}{
    \begin{tabular}{@{}l|cc|ccc@{}}

    \toprule
    \multicolumn{1}{@{}l|}{Model} & LLM & {Lane} & minADE & minFDE & MR \\ \midrule
    % \hline
    w/o LLM& -& $\checkmark$ &              1.30&   2.66& 0.47\\
    w/o LoRA& $\checkmark$& -&              1.27&   2.54& 0.43\\    
    w/o Lane& $\checkmark$& -&              1.32&   2.64& 0.49\\
    Traj-LLM& $\checkmark$& $\checkmark$&   1.24&   2.46& 0.41\\
    \bottomrule
    \end{tabular}}

    \label{tab: ablation result nuscene5}
\end{table}

\begin{table}[htpb]
\renewcommand\arraystretch{1.2}
    \centering
    \caption{\centering The result of the Traj-LLM and its variants in the ablation experiment for $K$=10.}
    \setlength{\tabcolsep}{3.5mm}{
    \begin{tabular}{@{}l|cc|ccc@{}}

    \toprule
    \multicolumn{1}{@{}l|}{Model} & LLM & {Lane} & minADE & minFDE & MR \\ \midrule
    % \hline
    w/o LLM& -& $\checkmark$ &              1.05&   1.91&   0.27\\
    w/o LoRA& $\checkmark$& -&              1.02&   1.82&   0.26\\    
    w/o Lane& $\checkmark$& -&              1.01&   1.75&   0.25\\
    Traj-LLM& $\checkmark$& $\checkmark$&   0.99&   1.73&   0.23\\
    \bottomrule
    \end{tabular}}

    \label{tab: ablation result nuscene10}
\end{table}

\subsection{Few-shot Study}
In the realm of trajectory prediction, the utilization of a select few datasets to reach exceptional performance is often a coveted ideal, owing to the exorbitant costs associated with data collection and maintenance. This few-shot scenario parallels real-world situations. To effectively evaluate the advantages of leveraging pre-trained LLMs in few-shot learning settings for trajectory prediction, we elaborately orchestrated a series of experiments.

\begin{figure}
  \centering
    \includegraphics[width=3.3in]{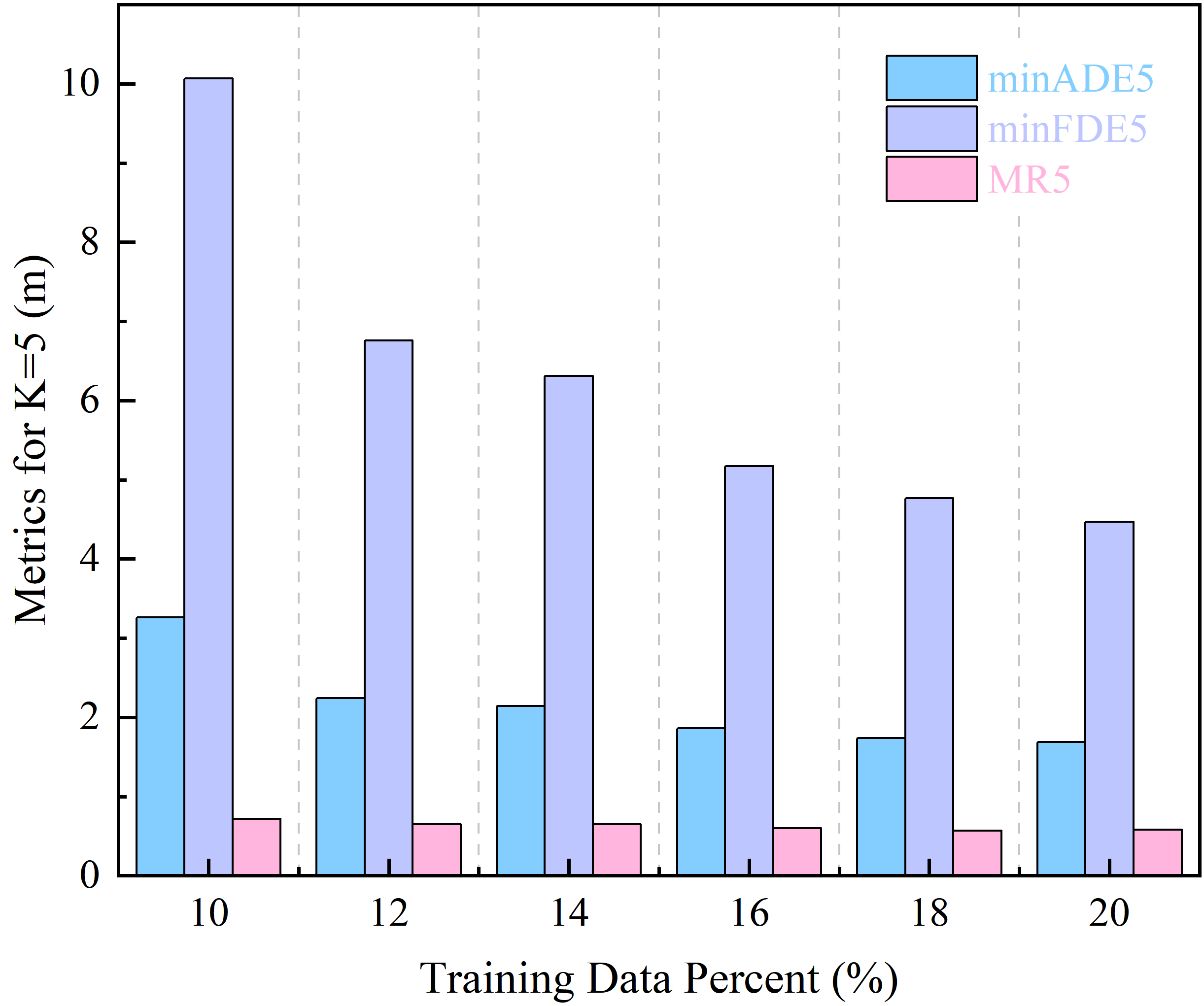}
    \caption{The result of the few-shot study for various metrics across training data percentages ranging from 10\% to 20\% under $K$=5.}
    \label{fig: few shot mr5}
\end{figure}

\begin{figure}
  \centering
    \includegraphics[width=3.3in]{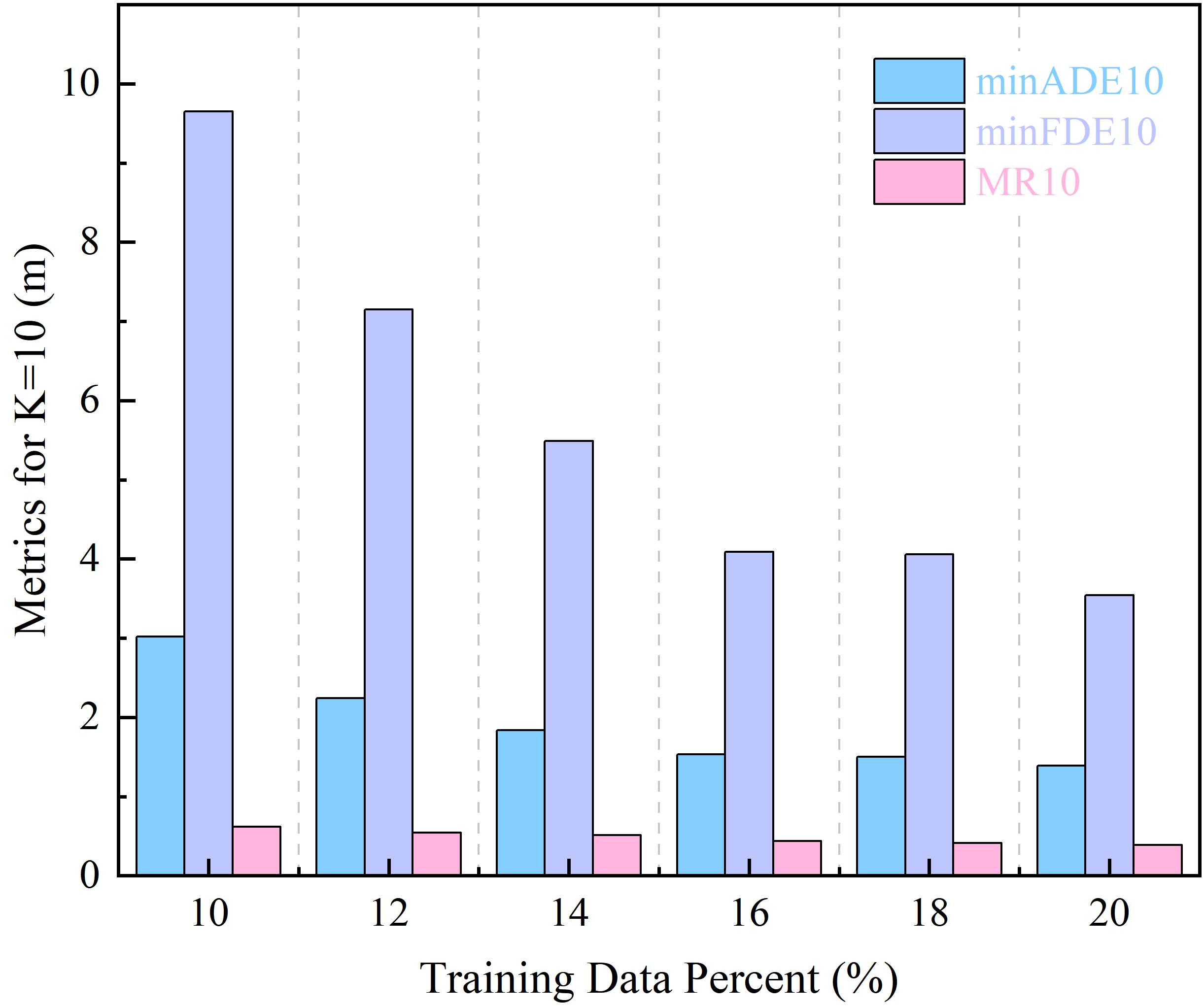}
    \caption{The result of the few-shot study for various metrics across training data percentages ranging from 10\% to 20\% under $K$=10.}
    \label{fig: few shot mr10}
\end{figure}

In these few-shot trials, we maintained consistent evaluation sets in line with the full-sample experiments, deliberately limiting the percentage of training data. Tables \ref{tab: few shot nuscene5} and \ref{tab: few shot nuscene10} present the outcomes of using only 10\% to 50\% of the training data for trajectory forecasting. Additionally, Fig. \ref{fig: few shot mr5} and Fig. \ref{fig: few shot mr10} illustrate various metrics across training data percentages ranging from 10\% to 20\%. Noteworthy is the revelation that Traj-LLM, even with only 50\% of the data, outperforms a majority of baselines reliant on 100\% data utilization across all evaluated metrics. Thanks to the innate representation learning capability encapsulated in LLMs and well-designed lane-aware probability learning, Traj-LLM consistently keeps commendable performance even in extremely sparse data scenarios (e.g., 10\% to 30\% training data). These compelling revelations stress Traj-LLM's remarkable capacity for generalization and adaptation, even when confronted with insufficient samples. 

% Despite experiencing a reduction of 33.3\% and 13.5\% in $\text{minADE}_5$ and $\text{minFDE}_5$ under the 10\% few-shot setting compared to the full-sample model, Traj-LLM still outshines the baseline model Trajectron++. 
% What's more, these experiments also confirm the profound effectiveness of pre-trained LLMs' few-shot learning capability within the trajectory prediction domain. 

\begin{table}[]
\renewcommand\arraystretch{1.2}
    \centering
    \caption{\centering The result of the few-shot study for $K$=5.}
    \setlength{\tabcolsep}{7.8mm}{
    \begin{tabular}{@{}l|ccc@{}}
    \toprule
    \multicolumn{1}{@{}l|}{Few-shot ratio} & minADE & minFDE & MR \\ \midrule
    % \hline   
    10\% &              3.26&   10.07&  0.72\\ % 30epoch 3.79 11.41 0.75 40epoch 3.41 10.52 0.72 50epoch 3.26 10.07 0.72
    20\% &              1.69&   4.47&   0.58\\ % 30epoch 1.94 5.20 0.58  40epoch 1.80 4.78 0.59  50epoch 1.69 4.47 0.58
    30\% &              1.42&   3.18&   0.51\\ % 30epoch 1.58 3.51 0.52  40epoch 1.46 3.28 0.53  50epoch 1.42 3.18 0.51
    40\% &              1.36&   2.94&   0.49\\ % 30epoch 1.50 3.35 0.55  40epoch 1.37 3.01 0.50  50epoch 1.36 2.94 0.49
    50\% &              1.30&   2.69&   0.44\\ % 30epoch 1.35 2.87 0.47  40epoch 1.31 2.73 0.45  50epoch 1.30 2.69 0.44
    \bottomrule
    \end{tabular}}

    \label{tab: few shot nuscene5}
\end{table}

% \begin{table}[htpb]
% \renewcommand\arraystretch{1.2}
%     \centering
%     \caption{\centering The result of the few-shot study for $K$=5.}
%     \setlength{\tabcolsep}{7.8mm}{
%     \begin{tabular}{@{}l|ccc@{}}
%     \toprule
%     \multicolumn{1}{@{}l|}{Few-shot ratio} & minADE & minFDE & MR \\ \midrule
%     % \hline 
%     10\% &              3.26&   10.07&   0.72\\ 50epoch √
%     12\% &              2.24&    6.76&   0.65\\ 50epoch √
%     14\% &              2.14&    6.31&   0.65\\ 50epoch √
%     16\% &              1.86&    5.17&   0.60\\ 50epoch √
%     18\% &              1.74&    4.77&   0.57\\ 50epoch √
%     20\% &              1.69&    4.47&   0.58\\ 50epoch √
%     \bottomrule
%     \end{tabular}}

%     \label{tab: few shot nuscene5}
% \end{table}

\begin{table}[]
\renewcommand\arraystretch{1.2}
    \centering
    \caption{\centering The result of the few-shot study for $K$=10.}
    \setlength{\tabcolsep}{7.8mm}{
    \begin{tabular}{@{}l|ccc@{}}
    \toprule
    \multicolumn{1}{@{}l|}{Few-shot ratio} & minADE & minFDE & MR \\ \midrule
    % \hline
    10\% &              3.02&   9.65&   0.62\\ % 30epoch 3.52 10.89 0.66 40epoch 3.20 10.14 0.65 50epoch 3.02 9.65 0.62
    20\% &              1.39&   3.54&   0.39\\ % 30epoch 1.50 3.93 0.42  40epoch 1.41 3.55 0.39  50epoch 1.39 3.54 0.39    
    30\% &              1.12&   2.26&   0.31\\ % 30epoch 1.21 2.51 0.33  40epoch 1.14 2.31 0.31  50epoch 1.12 2.26 0.31
    40\% &              1.05&   1.99&   0.27\\ % 30epoch 1.15 2.30 0.34  40epoch 1.07 2.04 0.28  50epoch 1.05 1.99 0.27
    50\% &              1.02&   1.85&   0.25\\ % 30epoch 1.06 1.96 0.28  40epoch 1.03 1.91 0.27  50epoch 1.02 1.85 0.25
    \bottomrule
    \end{tabular}}

    \label{tab: few shot nuscene10}
\end{table}

% \begin{table}[htpb]
% \renewcommand\arraystretch{1.2}
%     \centering
%     \caption{\centering The result of the few-shot study for $K$=10.}
%     \setlength{\tabcolsep}{7.8mm}{
%     \begin{tabular}{@{}l|ccc@{}}
%     \toprule
%     \multicolumn{1}{@{}l|}{Few-shot ratio} & minADE & minFDE & MR \\ \midrule
%     % \hline
%     10\% &              3.02&   9.65&   0.62\\ % 50epoch
%     12\% &              2.24&   7.15&   0.54\\ % 50epoch
%     14\% &              1.84&   5.49&   0.51\\ % 50epoch
%     16\% &              1.53&   4.09&   0.44\\ % 50epoch
%     18\% &              1.50&   4.06&   0.41\\ % 50epoch
%     20\% &              1.39&   3.54&   0.39\\ % 50epoch
%     \bottomrule
%     \end{tabular}}

%     \label{tab: few shot nuscene10}
% \end{table}

\subsection{Qualitative Analysis}
\subsubsection{Visualization of motion prediction under full-sample training} 
\begin{figure*}[]
  \centering
  \includegraphics[width=7.0in]{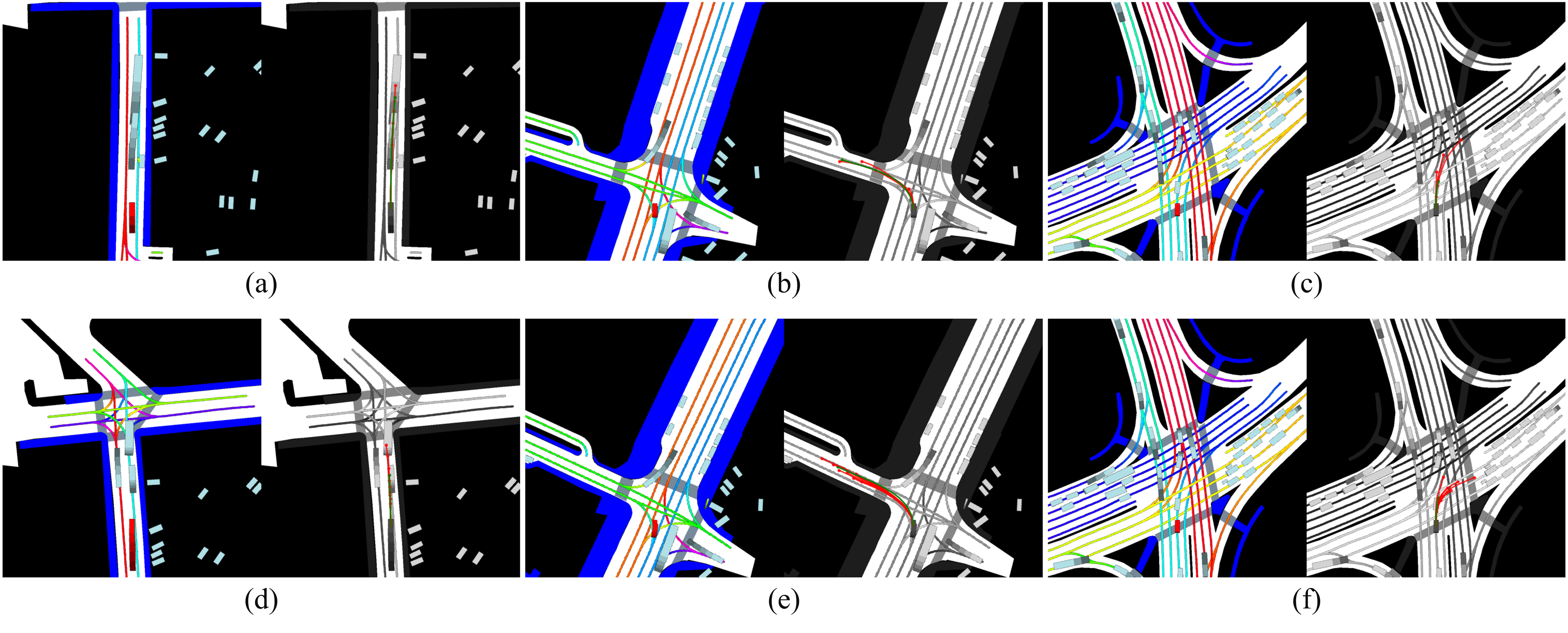}
    \caption{Qualitative results of Traj-LLM on various scenarios under full-sample training. In each subfigure, left: HD map, right: predictions and ground truth. Ground truth future trajectory is depicted in green lines, the predicted trajectories are in red lines. The top and bottom parts respectively exhibit the instances with prediction modalities $K$=5 and $K$=10.}
    \label{fig: nuscene visaul}
\end{figure*}

Illustrated in Fig. \ref{fig: nuscene visaul}, are the visualizations for predicted trajectories across different driving scenes in the full training dataset, enabling intuitive analysis of trajectory precision and diversity. The top and bottom parts of Fig. \ref{fig: nuscene visaul} respectively exhibit the instances under the conditions of $K$=5 and $K$=10. From these visual insights, we can find that whether maintaining a straight path or navigating intersections to execute left or right turns, the ground truth of the target agent closely corresponds with one of the multi-modal trajectories forecast by Traj-LLM in terms of trends. Furthermore, we observe that the trajectories generated by Traj-LLM display sound rationality and compliance, as they do not exceed the boundaries of the roadway. These comprehensive predictions concerning both lateral and longitudinal vehicular behaviors robustly prove the strong predictive power of Traj-LLM. 

\subsubsection{Visualization of Motion Prediction under few-shot settings with 50\% data samples} 
\begin{figure*}[]
  \centering
  \includegraphics[width=7.0in]{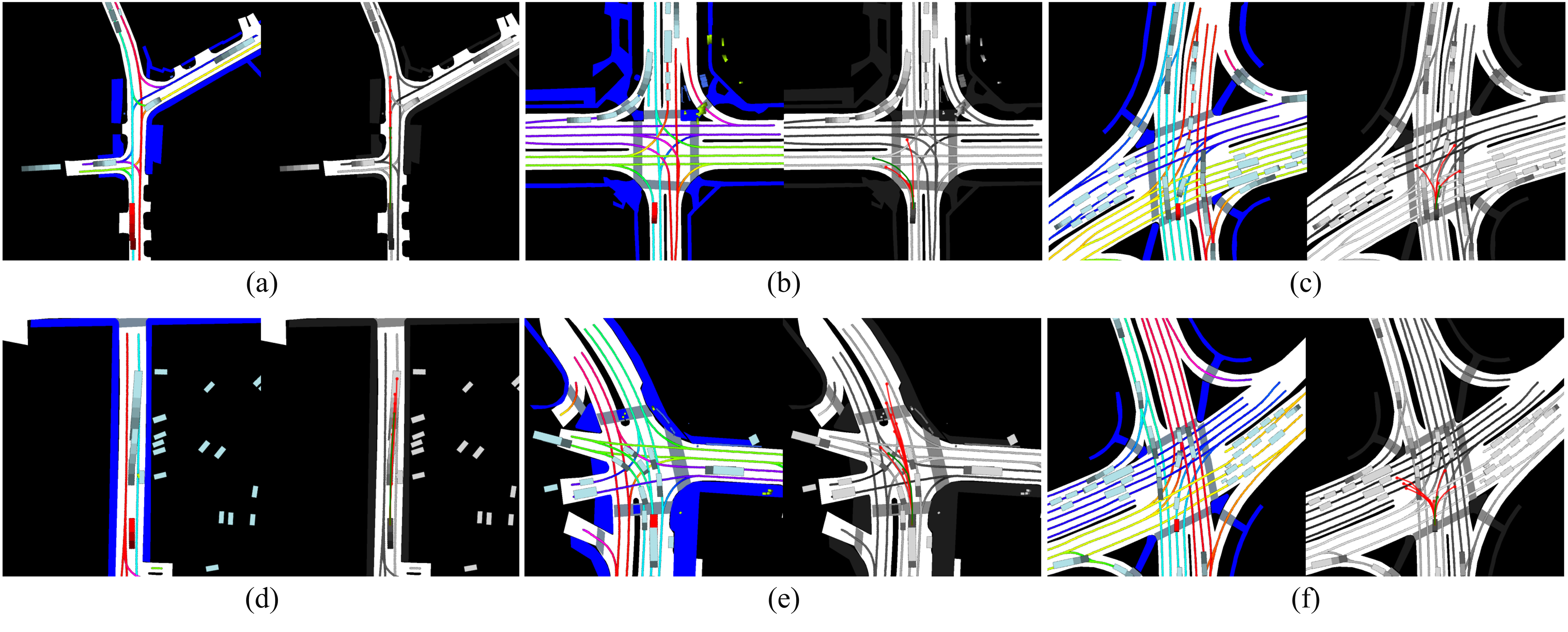}
    \caption{Qualitative results of Traj-LLM on various scenarios under few-shot settings with 50\% data samples. In each subfigure, left: HD map, right: predictions and ground truth. Ground truth future trajectory is shown in green lines, the predicted trajectories are in red lines. The top and bottom parts respectively exhibit the instances with prediction modalities $K$=5 and $K$=10.}
    \label{fig: nuscene visaul few shot}
\end{figure*}

Fig. \ref{fig: nuscene visaul few shot} showcases qualitative results of Traj-LLM in few-shot scenarios with $K$=5 and $K$=10. Despite being trained on only 50\% of the data, Traj-LLM consistently produces plausible predictions for the target agent across diverse scenarios, including intersections and straight driving. Furthermore, Traj-LLM accurately captures nuanced behaviors such as acceleration and deceleration. The uncertain feature of the agent's motion is also depicted with robust multi-modal predictions. These few-shot predictions further underscore the potent predictive capability of Traj-LLM, leveraging LLMs' exceptional scene understanding capacity and elaborately crafted lane-aware probability learning.

\section{Conclusions}
In this study, we propose the Traj-LLM, a novel trajectory prediction framework, that seeks to investigate the feasibility of LLMs in inferring agents' future trajectories without explicit prompt engineering design. Traj-LLM starts with sparse context joint coding, responsible for parsing the features of agents and scenes into a form that is understandable by LLMs. We then guide LLMs to learn a spectrum of high-level knowledge inherent in trajectory prediction tasks, such as scene context and social interactions. Aiming to imitate the human-like lane focus cognitive function and further enhance the scene understanding of Traj-LLM, we propose lane-aware probabilistic learning powered by the pioneering Mamba module. For achieving scene-compliant multi-modal predictions, this study introduces the multi-modal Laplace decoder. The experiments demonstrate that Traj-LLM outperforms state-of-the-art methods in various metrics and the few-shot study further validates Traj-LLM's outstanding prowess.

\bibliographystyle{IEEEtran}
\bibliography{ref}

\begin{thebibliography}{10}
\providecommand{\url}[1]{#1}
\csname url@samestyle\endcsname
\providecommand{\newblock}{\relax}
\providecommand{\bibinfo}[2]{#2}
\providecommand{\BIBentrySTDinterwordspacing}{\spaceskip=0pt\relax}
\providecommand{\BIBentryALTinterwordstretchfactor}{4}
\providecommand{\BIBentryALTinterwordspacing}{\spaceskip=\fontdimen2\font plus
\BIBentryALTinterwordstretchfactor\fontdimen3\font minus \fontdimen4\font\relax}
\providecommand{\BIBforeignlanguage}[2]{{%
\expandafter\ifx\csname l@#1\endcsname\relax
\typeout{** WARNING: IEEEtran.bst: No hyphenation pattern has been}%
\typeout{** loaded for the language `#1'. Using the pattern for}%
\typeout{** the default language instead.}%
\else
\language=\csname l@#1\endcsname
\fi
#2}}
\providecommand{\BIBdecl}{\relax}
\BIBdecl

\bibitem{wu2023multi}
Y.~Wu, L.~Wang, S.~Zhou, J.~Duan, G.~Hua, and W.~Tang, ``Multi-stream representation learning for pedestrian trajectory prediction,'' in \emph{Proceedings of the AAAI Conference on Artificial Intelligence}, vol.~37, no.~3, 2023, pp. 2875--2882.

\bibitem{mo2021graph}
X.~Mo, Y.~Xing, and C.~Lv, ``Graph and recurrent neural network-based vehicle trajectory prediction for highway driving,'' in \emph{2021 IEEE International Intelligent Transportation Systems Conference (ITSC)}.\hskip 1em plus 0.5em minus 0.4em\relax IEEE, 2021, pp. 1934--1939.

\bibitem{park2018sequence}
S.~H. Park, B.~Kim, C.~M. Kang, C.~C. Chung, and J.~W. Choi, ``Sequence-to-sequence prediction of vehicle trajectory via lstm encoder-decoder architecture,'' in \emph{2018 IEEE intelligent vehicles symposium (IV)}.\hskip 1em plus 0.5em minus 0.4em\relax IEEE, 2018, pp. 1672--1678.

\bibitem{liao2024bat}
H.~Liao, Z.~Li, H.~Shen, W.~Zeng, D.~Liao, G.~Li, and C.~Xu, ``Bat: Behavior-aware human-like trajectory prediction for autonomous driving,'' in \emph{Proceedings of the AAAI Conference on Artificial Intelligence}, vol.~38, no.~9, 2024, pp. 10\,332--10\,340.

\bibitem{sheng2022graph}
Z.~Sheng, Y.~Xu, S.~Xue, and D.~Li, ``Graph-based spatial-temporal convolutional network for vehicle trajectory prediction in autonomous driving,'' \emph{IEEE Transactions on Intelligent Transportation Systems}, vol.~23, no.~10, pp. 17\,654--17\,665, 2022.

\bibitem{liao2024physics}
H.~Liao, C.~Wang, Z.~Li, Y.~Li, B.~Wang, G.~Li, and C.~Xu, ``Physics-informed trajectory prediction for autonomous driving under missing observation,'' \emph{Available at SSRN 4809575}, 2024.

\bibitem{geng2023dynamic}
M.~Geng, Y.~Chen, Y.~Xia, and X.~M. Chen, ``Dynamic-learning spatial-temporal transformer network for vehicular trajectory prediction at urban intersections,'' \emph{Transportation research part C: emerging technologies}, vol. 156, p. 104330, 2023.

\bibitem{zuo2023trajectory}
Z.~Zuo, X.~Wang, S.~Guo, Z.~Liu, Z.~Li, and Y.~Wang, ``Trajectory prediction network of autonomous vehicles with fusion of historical interactive features,'' \emph{IEEE Transactions on Intelligent Vehicles}, 2023.

\bibitem{ren2024emsin}
Y.~Ren, Z.~Lan, L.~Liu, and H.~Yu, ``Emsin: Enhanced multi-stream interaction network for vehicle trajectory prediction,'' \emph{IEEE Transactions on Fuzzy Systems}, 2024.

\bibitem{liu2021kg}
Y.~Liu, Y.~Wan, L.~He, H.~Peng, and S.~Y. Philip, ``Kg-bart: Knowledge graph-augmented bart for generative commonsense reasoning,'' in \emph{Proceedings of the AAAI conference on artificial intelligence}, vol.~35, no.~7, 2021, pp. 6418--6425.

\bibitem{ouyang2022training}
L.~Ouyang, J.~Wu, X.~Jiang, D.~Almeida, C.~L. Wainwright, P.~Mishkin, C.~Zhang, S.~Agarwal, K.~Slama, A.~Ray \emph{et~al.}, ``Training language models to follow instructions with human feedback, 2022,'' \emph{URL https://arxiv. org/abs/2203.02155}, vol.~13, p.~1, 2022.

\bibitem{chen2023driving}
L.~Chen, O.~Sinavski, J.~H{\"u}nermann, A.~Karnsund, A.~J. Willmott, D.~Birch, D.~Maund, and J.~Shotton, ``Driving with llms: Fusing object-level vector modality for explainable autonomous driving,'' \emph{arXiv preprint arXiv:2310.01957}, 2023.

\bibitem{wu2023language}
D.~Wu, W.~Han, T.~Wang, Y.~Liu, X.~Zhang, and J.~Shen, ``Language prompt for autonomous driving,'' \emph{arXiv preprint arXiv:2309.04379}, 2023.

\bibitem{wen2023dilu}
L.~Wen, D.~Fu, X.~Li, X.~Cai, T.~Ma, P.~Cai, M.~Dou, B.~Shi, L.~He, and Y.~Qiao, ``Dilu: A knowledge-driven approach to autonomous driving with large language models,'' \emph{arXiv preprint arXiv:2309.16292}, 2023.

\bibitem{bae2024can}
I.~Bae, J.~Lee, and H.-G. Jeon, ``Can language beat numerical regression? language-based multimodal trajectory prediction,'' \emph{arXiv preprint arXiv:2403.18447}, 2024.

\bibitem{chib2024lg}
P.~S. Chib and P.~Singh, ``Lg-traj: Llm guided pedestrian trajectory prediction,'' \emph{arXiv preprint arXiv:2403.08032}, 2024.

\bibitem{xue2023promptcast}
H.~Xue and F.~D. Salim, ``Promptcast: A new prompt-based learning paradigm for time series forecasting,'' \emph{IEEE Transactions on Knowledge and Data Engineering}, 2023.

\bibitem{radford2021learning}
A.~Radford, J.~W. Kim, C.~Hallacy, A.~Ramesh, G.~Goh, S.~Agarwal, G.~Sastry, A.~Askell, P.~Mishkin, J.~Clark \emph{et~al.}, ``Learning transferable visual models from natural language supervision,'' in \emph{International conference on machine learning}.\hskip 1em plus 0.5em minus 0.4em\relax PMLR, 2021, pp. 8748--8763.

\bibitem{deo2018convolutional}
N.~Deo and M.~M. Trivedi, ``Convolutional social pooling for vehicle trajectory prediction,'' in \emph{Proceedings of the IEEE conference on computer vision and pattern recognition workshops}, 2018, pp. 1468--1476.

\bibitem{liang2021nettraj}
Y.~Liang and Z.~Zhao, ``Nettraj: A network-based vehicle trajectory prediction model with directional representation and spatiotemporal attention mechanisms,'' \emph{IEEE Transactions on Intelligent Transportation Systems}, vol.~23, no.~9, pp. 14\,470--14\,481, 2021.

\bibitem{wang2023spatio}
Z.~Wang, J.~Zhang, J.~Chen, and H.~Zhang, ``Spatio-temporal context graph transformer design for map-free multi-agent trajectory prediction,'' \emph{IEEE Transactions on Intelligent Vehicles}, 2023.

\bibitem{du2024social}
Q.~Du, X.~Wang, S.~Yin, L.~Li, and H.~Ning, ``Social force embedded mixed graph convolutional network for multi-class trajectory prediction,'' \emph{IEEE Transactions on Intelligent Vehicles}, 2024.

\bibitem{wu2023graph}
K.~Wu, Y.~Zhou, H.~Shi, X.~Li, and B.~Ran, ``Graph-based interaction-aware multimodal 2d vehicle trajectory prediction using diffusion graph convolutional networks,'' \emph{IEEE Transactions on Intelligent Vehicles}, 2023.

\bibitem{guo2023vectorized}
L.~Guo, C.~Shan, T.~Shi, X.~Li, and F.-Y. Wang, ``A vectorized representation model for trajectory prediction of intelligent vehicles in challenging scenarios,'' \emph{IEEE Transactions on Intelligent Vehicles}, 2023.

\bibitem{zhang2022ai}
K.~Zhang, L.~Zhao, C.~Dong, L.~Wu, and L.~Zheng, ``Ai-tp: Attention-based interaction-aware trajectory prediction for autonomous driving,'' \emph{IEEE Transactions on Intelligent Vehicles}, vol.~8, no.~1, pp. 73--83, 2022.

\bibitem{mo2023map}
X.~Mo, Y.~Xing, H.~Liu, and C.~Lv, ``Map-adaptive multimodal trajectory prediction using hierarchical graph neural networks,'' \emph{IEEE Robotics and Automation Letters}, 2023.

\bibitem{liao2024cognitive}
H.~Liao, Y.~Li, Z.~Li, C.~Wang, Z.~Cui, S.~E. Li, and C.~Xu, ``A cognitive-based trajectory prediction approach for autonomous driving,'' \emph{arXiv preprint arXiv:2402.19251}, 2024.

\bibitem{geng2023physics}
M.~Geng, J.~Li, Y.~Xia, and X.~M. Chen, ``A physics-informed transformer model for vehicle trajectory prediction on highways,'' \emph{Transportation research part C: emerging technologies}, vol. 154, p. 104272, 2023.

\bibitem{hu2023holistic}
H.~Hu, Q.~Wang, Z.~Zhang, Z.~Li, and Z.~Gao, ``Holistic transformer: A joint neural network for trajectory prediction and decision-making of autonomous vehicles,'' \emph{Pattern Recognition}, vol. 141, p. 109592, 2023.

\bibitem{roy2019vehicle}
D.~Roy, T.~Ishizaka, C.~K. Mohan, and A.~Fukuda, ``Vehicle trajectory prediction at intersections using interaction based generative adversarial networks,'' in \emph{2019 IEEE Intelligent transportation systems conference (ITSC)}.\hskip 1em plus 0.5em minus 0.4em\relax IEEE, 2019, pp. 2318--2323.

\bibitem{li2021vehicle}
X.~Li, G.~Rosman, I.~Gilitschenski, C.-I. Vasile, J.~A. DeCastro, S.~Karaman, and D.~Rus, ``Vehicle trajectory prediction using generative adversarial network with temporal logic syntax tree features,'' \emph{IEEE Robotics and Automation Letters}, vol.~6, no.~2, pp. 3459--3466, 2021.

\bibitem{neumeier2021variational}
M.~Neumeier, M.~Botsch, A.~Tollk{\"u}hn, and T.~Berberich, ``Variational autoencoder-based vehicle trajectory prediction with an interpretable latent space,'' in \emph{2021 IEEE International Intelligent Transportation Systems Conference (ITSC)}.\hskip 1em plus 0.5em minus 0.4em\relax IEEE, 2021, pp. 820--827.

\bibitem{chen2021trajvae}
X.~Chen, J.~Xu, R.~Zhou, W.~Chen, J.~Fang, and C.~Liu, ``Trajvae: A variational autoencoder model for trajectory generation,'' \emph{Neurocomputing}, vol. 428, pp. 332--339, 2021.

\bibitem{feng2019vehicle}
X.~Feng, Z.~Cen, J.~Hu, and Y.~Zhang, ``Vehicle trajectory prediction using intention-based conditional variational autoencoder,'' in \emph{2019 IEEE Intelligent Transportation Systems Conference (ITSC)}.\hskip 1em plus 0.5em minus 0.4em\relax IEEE, 2019, pp. 3514--3519.

\bibitem{ghoul2022lightweight}
A.~Ghoul, K.~Messaoud, I.~Yahiaoui, A.~Verroust-Blondet, and F.~Nashashibi, ``A lightweight goal-based model for trajectory prediction,'' in \emph{2022 IEEE 25th International Conference on Intelligent Transportation Systems (ITSC)}.\hskip 1em plus 0.5em minus 0.4em\relax IEEE, 2022, pp. 4209--4214.

\bibitem{yao2023goal}
Z.~Yao, X.~Li, B.~Lang, and M.~C. Chuah, ``Goal-lbp: Goal-based local behavior guided trajectory prediction for autonomous driving,'' \emph{IEEE Transactions on Intelligent Transportation Systems}, 2023.

\bibitem{dong2021multi}
B.~Dong, H.~Liu, Y.~Bai, J.~Lin, Z.~Xu, X.~Xu, and Q.~Kong, ``Multi-modal trajectory prediction for autonomous driving with semantic map and dynamic graph attention network,'' \emph{arXiv preprint arXiv:2103.16273}, 2021.

\bibitem{zhou2023query}
Z.~Zhou, J.~Wang, Y.-H. Li, and Y.-K. Huang, ``Query-centric trajectory prediction,'' in \emph{Proceedings of the IEEE/CVF Conference on Computer Vision and Pattern Recognition}, 2023, pp. 17\,863--17\,873.

\bibitem{jin2023time}
M.~Jin, S.~Wang, L.~Ma, Z.~Chu, J.~Y. Zhang, X.~Shi, P.-Y. Chen, Y.~Liang, Y.-F. Li, S.~Pan \emph{et~al.}, ``Time-llm: Time series forecasting by reprogramming large language models,'' \emph{arXiv preprint arXiv:2310.01728}, 2023.

\bibitem{chang2023llm4ts}
C.~Chang, W.-C. Peng, and T.-F. Chen, ``Llm4ts: Two-stage fine-tuning for time-series forecasting with pre-trained llms,'' \emph{arXiv preprint arXiv:2308.08469}, 2023.

\bibitem{gao2024units}
S.~Gao, T.~Koker, O.~Queen, T.~Hartvigsen, T.~Tsiligkaridis, and M.~Zitnik, ``Units: Building a unified time series model,'' \emph{arXiv preprint arXiv:2403.00131}, 2024.

\bibitem{bian2024multi}
Y.~Bian, X.~Ju, J.~Li, Z.~Xu, D.~Cheng, and Q.~Xu, ``Multi-patch prediction: Adapting llms for time series representation learning,'' \emph{arXiv preprint arXiv:2402.04852}, 2024.

\bibitem{gruver2024large}
N.~Gruver, M.~Finzi, S.~Qiu, and A.~G. Wilson, ``Large language models are zero-shot time series forecasters,'' \emph{Advances in Neural Information Processing Systems}, vol.~36, 2024.

\bibitem{yang2024human}
Y.~Yang, Q.~Zhang, C.~Li, D.~S. Marta, N.~Batool, and J.~Folkesson, ``Human-centric autonomous systems with llms for user command reasoning,'' in \emph{Proceedings of the IEEE/CVF Winter Conference on Applications of Computer Vision}, 2024, pp. 988--994.

\bibitem{cui2024drive}
C.~Cui, Y.~Ma, X.~Cao, W.~Ye, and Z.~Wang, ``Drive as you speak: Enabling human-like interaction with large language models in autonomous vehicles,'' in \emph{Proceedings of the IEEE/CVF Winter Conference on Applications of Computer Vision}, 2024, pp. 902--909.

\bibitem{gu2021densetnt}
J.~Gu, C.~Sun, and H.~Zhao, ``Densetnt: End-to-end trajectory prediction from dense goal sets,'' in \emph{Proceedings of the IEEE/CVF International Conference on Computer Vision}, 2021, pp. 15\,303--15\,312.

\bibitem{liu2023laformer}
M.~Liu, H.~Cheng, L.~Chen, H.~Broszio, J.~Li, R.~Zhao, M.~Sester, and M.~Y. Yang, ``Laformer: Trajectory prediction for autonomous driving with lane-aware scene constraints,'' \emph{arXiv preprint arXiv:2302.13933}, 2023.

\bibitem{dauphin2017language}
Y.~N. Dauphin, A.~Fan, M.~Auli, and D.~Grangier, ``Language modeling with gated convolutional networks,'' in \emph{International conference on machine learning}.\hskip 1em plus 0.5em minus 0.4em\relax PMLR, 2017, pp. 933--941.

\bibitem{zhang2023llama}
R.~Zhang, J.~Han, C.~Liu, A.~Zhou, P.~Lu, H.~Li, P.~Gao, and Y.~Qiao, ``Llama-adapter: Efficient fine-tuning of large language models with zero-initialized attention,'' in \emph{The Twelfth International Conference on Learning Representations}, 2023.

\bibitem{zhang2022tip}
R.~Zhang, W.~Zhang, R.~Fang, P.~Gao, K.~Li, J.~Dai, Y.~Qiao, and H.~Li, ``Tip-adapter: Training-free adaption of clip for few-shot classification,'' in \emph{European conference on computer vision}.\hskip 1em plus 0.5em minus 0.4em\relax Springer, 2022, pp. 493--510.

\bibitem{radford2019language}
A.~Radford, J.~Wu, R.~Child, D.~Luan, D.~Amodei, I.~Sutskever \emph{et~al.}, ``Language models are unsupervised multitask learners,'' \emph{OpenAI blog}, vol.~1, no.~8, p.~9, 2019.

\bibitem{hu2021lora}
E.~J. Hu, Y.~Shen, P.~Wallis, Z.~Allen-Zhu, Y.~Li, S.~Wang, L.~Wang, and W.~Chen, ``Lora: Low-rank adaptation of large language models,'' \emph{arXiv preprint arXiv:2106.09685}, 2021.

\bibitem{gu2023mamba}
A.~Gu and T.~Dao, ``Mamba: Linear-time sequence modeling with selective state spaces,'' \emph{arXiv preprint arXiv:2312.00752}, 2023.

\bibitem{elfwing2018sigmoid}
S.~Elfwing, E.~Uchibe, and K.~Doya, ``Sigmoid-weighted linear units for neural network function approximation in reinforcement learning,'' \emph{Neural networks}, vol. 107, pp. 3--11, 2018.

\bibitem{zhou2022hivt}
Z.~Zhou, L.~Ye, J.~Wang, K.~Wu, and K.~Lu, ``Hivt: Hierarchical vector transformer for multi-agent motion prediction,'' in \emph{Proceedings of the IEEE/CVF Conference on Computer Vision and Pattern Recognition}, 2022, pp. 8823--8833.

\bibitem{caesar2020nuscenes}
H.~Caesar, V.~Bankiti, A.~H. Lang, S.~Vora, V.~E. Liong, Q.~Xu, A.~Krishnan, Y.~Pan, G.~Baldan, and O.~Beijbom, ``nuscenes: A multimodal dataset for autonomous driving,'' in \emph{Proceedings of the IEEE/CVF conference on computer vision and pattern recognition}, 2020, pp. 11\,621--11\,631.

\bibitem{salzmann2020trajectron++}
T.~Salzmann, B.~Ivanovic, P.~Chakravarty, and M.~Pavone, ``Trajectron++: Dynamically-feasible trajectory forecasting with heterogeneous data,'' in \emph{Computer Vision--ECCV 2020: 16th European Conference, Glasgow, UK, August 23--28, 2020, Proceedings, Part XVIII 16}.\hskip 1em plus 0.5em minus 0.4em\relax Springer, 2020, pp. 683--700.

\bibitem{narayanan2021divide}
S.~Narayanan, R.~Moslemi, F.~Pittaluga, B.~Liu, and M.~Chandraker, ``Divide-and-conquer for lane-aware diverse trajectory prediction,'' in \emph{Proceedings of the IEEE/CVF Conference on Computer Vision and Pattern Recognition}, 2021, pp. 15\,799--15\,808.

\bibitem{cheng2023gatraj}
H.~Cheng, M.~Liu, L.~Chen, H.~Broszio, M.~Sester, and M.~Y. Yang, ``Gatraj: A graph-and attention-based multi-agent trajectory prediction model,'' \emph{ISPRS Journal of Photogrammetry and Remote Sensing}, vol. 205, pp. 163--175, 2023.

\bibitem{wang2022stepwise}
C.~Wang, Y.~Wang, M.~Xu, and D.~J. Crandall, ``Stepwise goal-driven networks for trajectory prediction,'' \emph{IEEE Robotics and Automation Letters}, vol.~7, no.~2, pp. 2716--2723, 2022.

\bibitem{khandelwal2020if}
S.~Khandelwal, W.~Qi, J.~Singh, A.~Hartnett, and D.~Ramanan, ``What-if motion prediction for autonomous driving,'' \emph{arXiv preprint arXiv:2008.10587}, 2020.

\bibitem{messaoud2021trajectory}
K.~Messaoud, N.~Deo, M.~M. Trivedi, and F.~Nashashibi, ``Trajectory prediction for autonomous driving based on multi-head attention with joint agent-map representation,'' in \emph{2021 IEEE Intelligent Vehicles Symposium (IV)}.\hskip 1em plus 0.5em minus 0.4em\relax IEEE, 2021, pp. 165--170.

\bibitem{yuan2021agentformer}
Y.~Yuan, X.~Weng, Y.~Ou, and K.~M. Kitani, ``Agentformer: Agent-aware transformers for socio-temporal multi-agent forecasting,'' in \emph{Proceedings of the IEEE/CVF International Conference on Computer Vision}, 2021, pp. 9813--9823.

\bibitem{kim2021lapred}
B.~Kim, S.~H. Park, S.~Lee, E.~Khoshimjonov, D.~Kum, J.~Kim, J.~S. Kim, and J.~W. Choi, ``Lapred: Lane-aware prediction of multi-modal future trajectories of dynamic agents,'' in \emph{Proceedings of the IEEE/CVF Conference on Computer Vision and Pattern Recognition}, 2021, pp. 14\,636--14\,645.

\bibitem{deo2020trajectory}
N.~Deo and M.~M. Trivedi, ``Trajectory forecasts in unknown environments conditioned on grid-based plans,'' \emph{arXiv preprint arXiv:2001.00735}, 2020.

\bibitem{gilles2022gohome}
T.~Gilles, S.~Sabatini, D.~Tsishkou, B.~Stanciulescu, and F.~Moutarde, ``Gohome: Graph-oriented heatmap output for future motion estimation,'' in \emph{2022 international conference on robotics and automation (ICRA)}.\hskip 1em plus 0.5em minus 0.4em\relax IEEE, 2022, pp. 9107--9114.

\bibitem{schafer2022context}
M.~Sch{\"a}fer, K.~Zhao, M.~B{\"u}hren, and A.~Kummert, ``Context-aware scene prediction network (caspnet),'' in \emph{2022 IEEE 25th International Conference on Intelligent Transportation Systems (ITSC)}.\hskip 1em plus 0.5em minus 0.4em\relax IEEE, 2022, pp. 3970--3977.

\bibitem{lee2022muse}
M.~Lee, S.~S. Sohn, S.~Moon, S.~Yoon, M.~Kapadia, and V.~Pavlovic, ``Muse-vae: Multi-scale vae for environment-aware long term trajectory prediction,'' in \emph{Proceedings of the IEEE/CVF Conference on Computer Vision and Pattern Recognition}, 2022, pp. 2221--2230.

\bibitem{girgis2021latent}
R.~Girgis, F.~Golemo, F.~Codevilla, M.~Weiss, J.~A. D'Souza, S.~E. Kahou, F.~Heide, and C.~Pal, ``Latent variable sequential set transformers for joint multi-agent motion prediction,'' \emph{arXiv preprint arXiv:2104.00563}, 2021.

\bibitem{gilles2021thomas}
T.~Gilles, S.~Sabatini, D.~Tsishkou, B.~Stanciulescu, and F.~Moutarde, ``Thomas: Trajectory heatmap output with learned multi-agent sampling,'' \emph{arXiv preprint arXiv:2110.06607}, 2021.

\bibitem{choi2022hierarchical}
D.~Choi and K.~Min, ``Hierarchical latent structure for multi-modal vehicle trajectory forecasting,'' in \emph{European Conference on Computer Vision}.\hskip 1em plus 0.5em minus 0.4em\relax Springer, 2022, pp. 129--145.

\bibitem{deo2022multimodal}
N.~Deo, E.~Wolff, and O.~Beijbom, ``Multimodal trajectory prediction conditioned on lane-graph traversals,'' in \emph{Conference on Robot Learning}.\hskip 1em plus 0.5em minus 0.4em\relax PMLR, 2022, pp. 203--212.

\end{thebibliography}

\vfill

\end{document}